\documentclass{article}
\usepackage{arxiv}
\usepackage[utf8]{inputenc} 
\usepackage[T1]{fontenc}    
\usepackage{hyperref}       
\usepackage{url}            
\usepackage{booktabs}       
\usepackage{amsfonts}       
\usepackage{nicefrac}       
\usepackage{microtype}      
\usepackage{lipsum}		
\usepackage{graphicx}
\usepackage{natbib}
\usepackage{doi}
\usepackage{array}
\usepackage{float}
\usepackage{color,soul}
\usepackage{amsmath}
\usepackage{caption}
\title{Domain-Adversarial Neural Network and Explainable AI for Reducing Tissue-of-Origin Signal in Pan-cancer Mortality Classification}

\author{%
\href{https://orcid.org/0000-0001-9956-3484}{\includegraphics[scale=0.06]{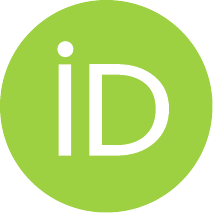}\hspace{1mm}Cristian Padron-Manrique} \\
Human Systems Biology Laboratory \\ 
Instituto Nacional de Medicina Genómica (INMEGEN) \\ 
Programa de Doctorado en Ciencias Biomédicas\\ Universidad Nacional Autónoma de México (UNAM).\\
Mexico City, Mexico\\
\texttt{inu\_juliocesar@hotmail.com} \\
\And
\href{https://orcid.org/0000-0003-1093-9970}{\includegraphics[scale=0.06]{orcid.pdf}\hspace{1mm}Juan Jos\'e Oropeza Valdez} \\
Centro de Ciencias de la Complejidad (C3) \\ 
Universidad Nacional Autónoma de México (UNAM) \\ 
Mexico City, Mexico\\
\texttt{juan.oropeza@c3.unam.mx} \\
\And
\href{https://orcid.org/0000-0001-5220-541X}{\includegraphics[scale=0.06]{orcid.pdf}\hspace{1mm}Osbaldo Resendis-Antonio} \\
Human Systems Biology Laboratory \\ 
Instituto Nacional de Medicina Genómica (INMEGEN) \\
Coordinación de la Investigación Científica\\
Red de Apoyo a la Investigación (RAI)\\
Centro de Ciencias de la Complejidad (C3)\\
Universidad Nacional Autónoma de México (UNAM) \\ 
Mexico City, Mexico\\
\texttt{oresendis@inmegen.gob.mx} \\
}

\renewcommand{\shorttitle}{DANN for Domain-Invariant Mortality Classification}

\hypersetup{
pdftitle={DOMAIN-ADVERSARIAL NEURAL NETWORK APPROACH FOR
REDUCING TISSUE-OF-ORIGIN SIGNAL IN EXPLAINABLE AI FOR
DOMAIN-INVARIANT MORTALITY CLASSIFICATION},
pdfsubject={DANN},
pdfauthor={Cristian Padron-Manrique, Juan Jos\'e Oropeza Valdez, Osbaldo Resendis-Antonio},
pdfkeywords={Cancer, Pancancer, Neural Networks, Deep Learning}
}

\begin{document}
\maketitle

\begin{abstract}
Tissue-of-origin signals dominate pan-cancer gene expression, often obscuring molecular features linked to patient survival. This hampers the discovery of generalizable biomarkers, as models tend to overfit tissue-specific patterns rather than capture survival-relevant signals. To address this, we propose a \textbf{Domain-Adversarial Neural Network (DANN)} trained on TCGA RNA-seq data to learn representations less biased by tissue and more focused on survival. Identifying tissue-independent genetic profiles is key to revealing core cancer programs. We assess the DANN using:
\begin{enumerate}
\item \textbf{Standard SHAP}, based on the original input space and DANN's mortality classifier.
\item A \textbf{layer-aware strategy} applied to hidden activations, including:
\begin{itemize}
\item An \textbf{unsupervised manifold} from \textbf{raw activations}.
\item A \textbf{supervised manifold} from \textit{mortality-specific SHAP values}.
\end{itemize}
\end{enumerate}
\textbf{Standard SHAP} remains confounded by tissue signals due to biases inherent in its computation. The \textbf{raw activation manifold} was dominated by high-magnitude activations, which masked subtle tissue and mortality-related signals. In contrast, the \textbf{layer-aware SHAP manifold} offers improved low-dimensional representations of both tissue and mortality signals, independent of activation strength, enabling subpopulation stratification and pan-cancer identification of survival-associated genes.

\end{abstract}


\section{Introduction}

Despite significant advances in cancer treatment, cancer remains a major global health challenge, ranking as the second leading cause of death and causing over 8 million deaths annually. Its incidence is projected to increase by more than 50\% in the coming decades \citep{RN1, RN2}. To address this growing burden, pan-cancer analyses seek to identify molecular features shared across diverse tumor types, aiming to uncover common druggable targets and develop therapies with broader clinical applicability \citep{RN3}.

The Cancer Genome Atlas (TCGA) provides a comprehensive multi-omics resource, including transcriptomic, genomic, epigenomic, and proteomic data from over 10,000 samples across 33 cancer types \citep{RN4}. A key objective of TCGA is to enable pan-cancer analysis, which focuses on identifying molecular mechanisms that transcend tissue boundaries. Unlike tumor-specific studies, this approach aims to reveal universal drivers of tumorigenesis and therapeutic opportunities independent of tissue context.

However, predicting survival outcomes—such as vital status, survival time, or mortality—remains a significant challenge in pan-cancer analysis. Gene expression profiles are heavily influenced by tissue-of-origin signals, which obscure molecular signatures consistent across tumor types. This confounding effect is evident in low-dimensional embeddings of TCGA data, where samples cluster strongly by tissue, reflecting tissue-specific expression patterns rather than shared pan-cancer features \citep{RN5, RN6}.

To mitigate tissue-of-origin bias in unsupervised feature selection, \citet{RN5} applied a linear residual orthogonal projection to gene expression data, removing variance attributable to tissue covariates. This method projects the data onto the orthogonal complement of the tissue-defined subspace, eliminating components linearly associated with tissue effects. However, transcriptomic data often exhibit complex, non-linear structures. Non-linear dimensionality reduction techniques, such as UMAP \citep{RN6}, reveal distinct tissue-driven clusters, suggesting that linear corrections may not entirely eliminate tissue-derived signals, potentially limiting the effectiveness and interpretability of downstream analyses.

Several studies have leveraged TCGA data to develop classification models across multiple cancer types \citep{RN4, Sun_2023}. However, these approaches typically do not account for tissue-of-origin confounding, which may restrict the generalizability of survival predictors across tumor types. Additionally, some studies have employed Explainable Artificial Intelligence (XAI) techniques, such as SHAP (SHapley Additive exPlanations), to identify features driving tissue-specific classification in TCGA’s 33 cancer types \citep{RN6, RN7}. While these methods highlight tissue-associated expression patterns, they do not evaluate the impact of removing tissue-origin signals \citep{RN6, RN7}.

In predictive modeling, models often overfit to dominant patterns unrelated to the target variable, particularly in the presence of class imbalance or strong confounding signals. This phenomenon is well-documented in image-based datasets, such as X-ray classification tasks, where models may exploit spurious correlations from hospital sources or imaging devices \citep{Kumari_2024}. Similarly, in TCGA transcriptomic data, cancer type—closely tied to tissue-of-origin—represents a significant source of variation \citep{RN5}. Moreover, survival labels are typically imbalanced, with more samples from living patients. This combination of dominant tissue signals and outcome imbalance increases the risk that models prioritize statistically prevalent patterns over biologically relevant ones, reinforcing spurious associations.

Domain adaptation techniques offer a solution by learning representations invariant to irrelevant sources of variation, such as tissue-of-origin, while preserving task-relevant signals. In this context, cancer types can be treated as distinct domains introducing structured variability in gene expression. One effective method is the Domain-Adversarial Neural Network (DANN) \citep{RN8}, which uses a gradient reversal layer to learn domain-invariant features by minimizing domain discrimination while optimizing target prediction. To address tissue-of-origin confounding in pan-cancer transcriptomic analysis, we propose a DANN architecture trained on TCGA RNA-seq data to learn representations less biased by tissue identity and more aligned with survival-relevant signals. Identifying tissue-independent genetic profiles is critical for uncovering core cancer programs that transcend organ-specific contexts.

We evaluated the interpretability and biological relevance of the DANN using two complementary approaches. First, we applied standard SHAP to the original gene expression space to explain the mortality classifier’s output. We then projected these SHAP values onto a UMAP embedding to visualize how the model distributed feature importance across samples in relation to survival. Second, we implemented a layer-aware strategy on hidden layer activations, deriving:  
(i) an unsupervised manifold based on raw activations, and  
(ii) a SHAP-based manifold computed from hidden layer activations for the mortality output.

Although the DANN effectively reduced tissue-of-origin signals through adversarial learning, \textbf{standard SHAP} with UMAP remained confounded by these effects. Since SHAP assigns importance directly to the input features—without leveraging the internal representations learned by the model—it operates on the original gene expression matrix, which is inherently biased by tissue identity. As a result, it continued to prioritize tissue-specific patterns, even when these were no longer relevant for classification. To assess whether the DANN aligned with the survival prediction task in a cancer-type invariant manner, we compared two representations: the \textbf{raw activation manifold} and the \textbf{layer-aware SHAP manifold}. The \textbf{raw activation manifold} was dominated by high-magnitude activations that obscured subtle mortality-related patterns. In contrast, the \textbf{layer-aware SHAP manifold} provided a clearer and more interpretable low-dimensional representation of the model’s decision-making process—independent of activation strength—and revealed subpopulations associated with distinct survival outcomes, not driven by tissue-of-origin.

These findings highlight the limitations of input-level explanations and underscore the value of analyzing hidden representations to verify whether neural networks learn task-relevant biological signals in pan-cancer settings.

\section{Results}
\label{sec:Results}
\subsection{Dominance of tissue origin of cancer for low dimensional manifolds}

\begin{figure}
    \centering
    \includegraphics[width=0.9\textwidth]{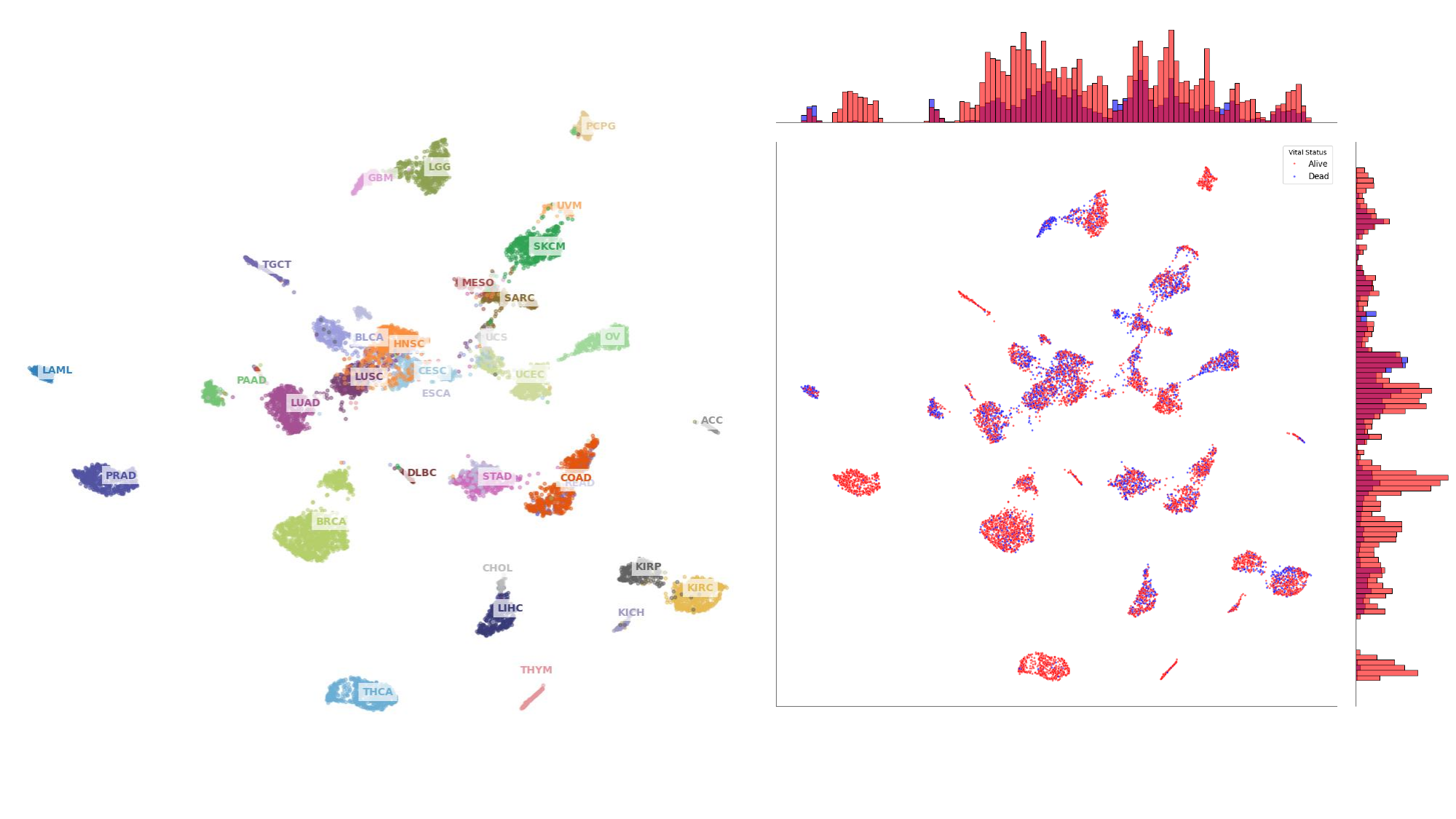}
    \caption{
    \textbf{Tissue-of-origin patterns dominate unsupervised transcriptomic projections.} Unsupervised methods such as UMAP separate tissue-of-origin clusters but fail to achieve similar separation for vital status. Unsupervised methods guide the separation of tissue-of-origin clusters (left), enabling UMAP to delineate cancer panels based on transcriptomic profiles. However, for vital status (right), UMAP fails to achieve meaningful class separation between alive (red) and dead (blue) samples, highlighting the challenge of separating survival status with unsupervised methods.
    }
    \label{fig:Figure1}
\end{figure}

Our analysis begins with the observation that transcriptomic profiles strongly shape the structure of TCGA RNA-seq data, as revealed by a UMAP embedding. As shown on the left side of Figure~\ref{fig:Figure1}, UMAP clusters the samples primarily by tissue of origin, illustrating how tissue-specific gene expression signatures dominate the data. This indicates that transcriptomic variation across tissues is the principal factor driving sample organization in an unsupervised setting.

However, when samples are colored by survival status (red for alive, blue for deceased), no clear separation is observed, as shown on the right side of the figure. This suggests that survival outcomes do not naturally structure the data in the same way as tissue identity. The inability of UMAP to distinguish between alive and deceased patients further highlights the overwhelming influence of tissue-of-origin signals, which obscure survival-related variation.

This observation reveals a key challenge: the survival signal is relatively weak and does not manifest as a dominant structure in the transcriptomic space. Furthermore, the data set is unbalanced for survival status, which can bias models toward learning the more separable tissue signal while neglecting subtler patterns associated with patient outcomes.

Unlike tissue identity—which produces clear, intrinsic structure in high-dimensional gene expression data—survival is likely influenced by a complex interplay of transcriptomic, genetic, and clinical variables that unsupervised techniques do not capture easily. These findings underscore the limitations of conventional dimensionality reduction methods such as UMAP in revealing biologically meaningful survival signals, and highlight the need for approaches that can disentangle survival-relevant information from confounding factors like tissue of origin.

\subsection{Domain-Adversarial Neural Network (DANN) Architecture Loss and Performance Evaluations}

\begin{figure}
    \centering
    \includegraphics[width=0.9\textwidth]{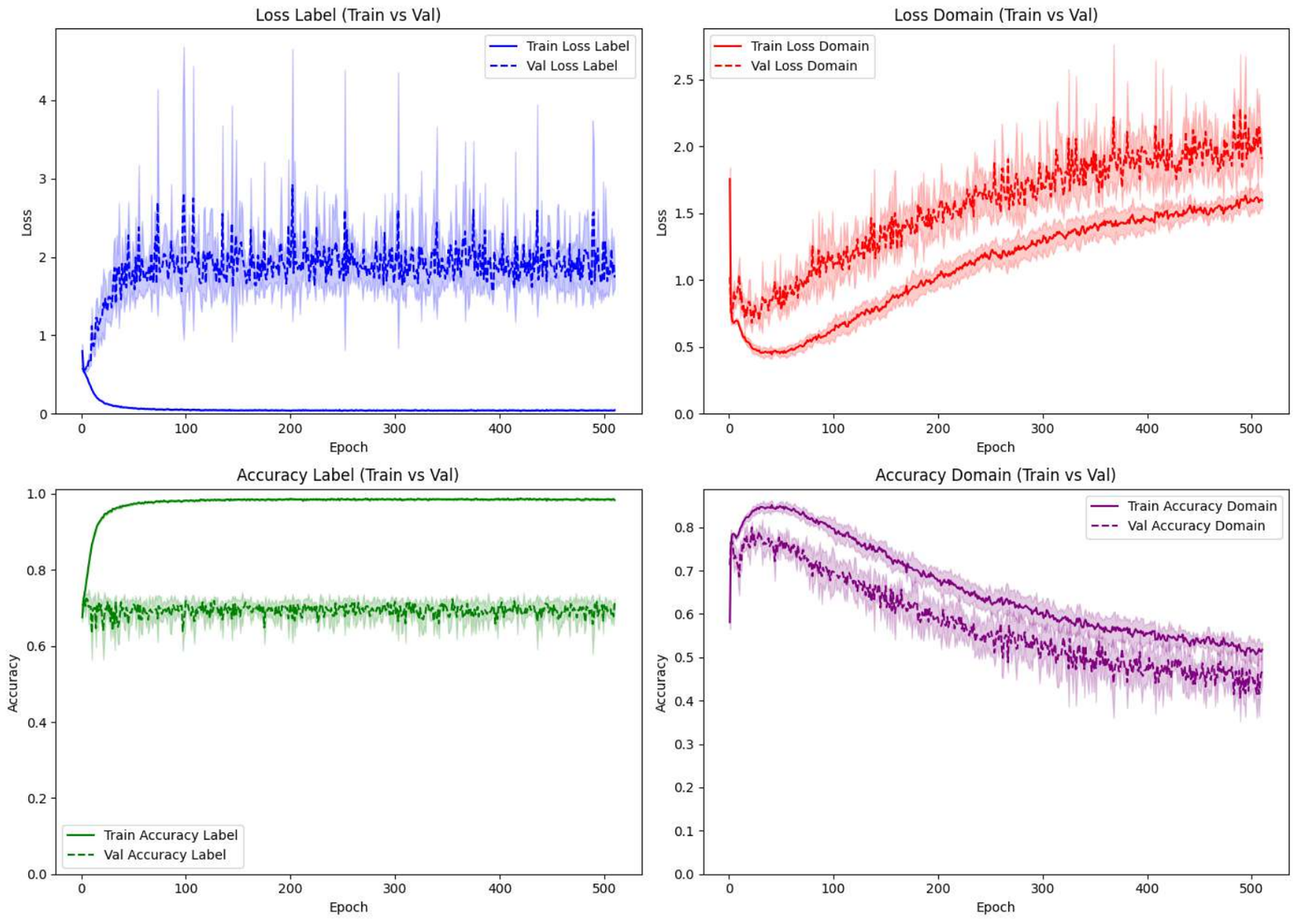}
    \caption{
    \textbf{Training and validation performance of the DANN model over 499 epochs.} This figure presents the Domain-Adversarial Neural Network (DANN) training and validation performance metrics across 499 epochs. Top-left: Training and cross-validation loss for the label classifier. Top-right: Training and cross-validation loss for the domain classifier. Bottom-left: Training and cross-validation accuracy for the label classifier. Bottom-right: Training and cross-validation accuracy for the domain classifier. The shaded regions in each plot represent the standard deviation across cross-validation folds.
    }
    \label{fig:Figure2}
\end{figure}

Training and validation performance metrics for the Domain-Adversarial Neural Network (DANN) over 499 epochs reveal distinct trends for the label and domain classifiers (Figure \ref{fig:Figure2}). The DANN architecture includes two classification components: the label classifier, which predicts the vital status of tumor samples—distinguishing between patients who are alive or deceased, thus directly addressing mortality—and the domain classifier, which aims to reduce the influence of tissue-of-origin biases by adversarially predicting the cancer type (domain). This setup allows the model to focus on mortality-relevant transcriptomic signals while minimizing confounding effects from cancer-type-specific expression patterns across the 33 TCGA cancer types (panels). In the label classifier loss (top-left), training loss decreases sharply in the early epochs. It stabilizes close to zero, suggesting that the model fits the training data well (average training loss: 0.0457). However, the validation loss remains relatively high and fluctuates significantly, averaging 1.7353, indicating potential challenges in generalizing across domains. Although the validation loss is high and noisy, the validation accuracy remains consistent (\~70\%), suggesting that the model generalizes reasonably well on the main task (label classification). The domain classifier loss (top-right) follows a different trajectory, decreasing in the initial epochs but gradually increasing over time. This suggests that the domain classifier struggles to maintain meaningful discrimination between domains as training progresses, likely due to the adversarial dynamics between the domain classifier and the feature extractor.

In the label classifier accuracy (bottom-left), training accuracy rapidly reaches near-perfect performance (average training accuracy: 98.35\%), whereas validation accuracy remains consistently lower at around 71.25\%. Rather than pointing solely to overfitting, this discrepancy may reflect the difficulty the label classifier has in achieving generalization, possibly due to the feature extractor being influenced by adversarial pressure from the domain classifier. As the feature extractor is optimized to learn domain-invariant representations, it may suppress domain-specific features that are also informative for label prediction, resulting in degraded performance on unseen domains. Finally, the domain classifier accuracy (bottom-right) initially increases but then steadily declines, indicating that the model effectively reduces domain discrimination over time—as expected in adversarial training. This trend suggests that the feature extractor is successfully learning domain-invariant representations.

Overall, these results highlight the effectiveness of DANN in aligning feature distributions across domains while revealing a trade-off in label classification performance. The increasing domain loss and decreasing domain accuracy are consistent with successful adversarial training. However, the performance gap between training and validation in the label classification task underscores the need to improve domain generalization without suppressing label-relevant features.

\subsection{DANN Architecture Clustering Evaluations}
\label{subsec:Clustering}
\begin{figure}
    \centering
    \includegraphics[width=0.9\textwidth]{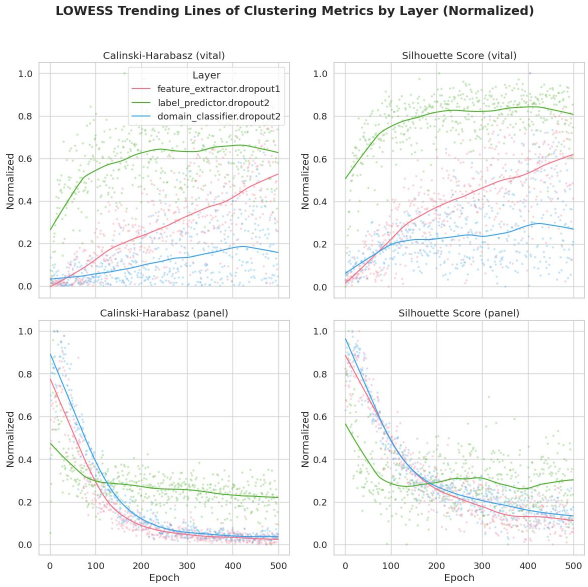}
    \caption{
    \textbf{Normalized clustering scores reveal domain and survival structure changes during training.} Normalized clustering scores (Calinski-Harabasz and Silhouette) computed on the 2D UMAP projections of layer activations from the DANN architecture over training epochs. Top panels reflect clustering quality for \textbf{vital status} (alive vs dead), while bottom panels show clustering based on \textbf{domain labels} (TCGA cancer types). Each line corresponds to a specific transformation layer within the model: \texttt{feature\_extractor.dropout1} (red), \texttt{label\_predictor.dropout2} (green), and \texttt{domain\_classifier.dropout2} (blue), which represent the last transformation in each respective path. LOWESS smoothing reveals a consistent decline in domain-related clustering—particularly pronounced in the label predictor pathway—while clustering by vital status becomes progressively stronger. This illustrates the model’s ability to suppress domain bias and enhance survival-relevant structure during training.
    }
    \label{fig:Figure3}
\end{figure}

We tracked how hidden representations evolved across different network layers throughout training to evaluate the effectiveness of domain-invariant feature learning in our DANN architecture. As shown in Figure \ref{fig:Figure3}, we computed the Calinski-Harabasz and Silhouette Scores—both normalized clustering metrics—on the 2D UMAP projections of three key layers: feature\_extractor.dropout1, label\_predictor.dropout2, and domain\_classifier.dropout2.

\begin{figure}
    \centering
    \includegraphics[width=0.9\textwidth]{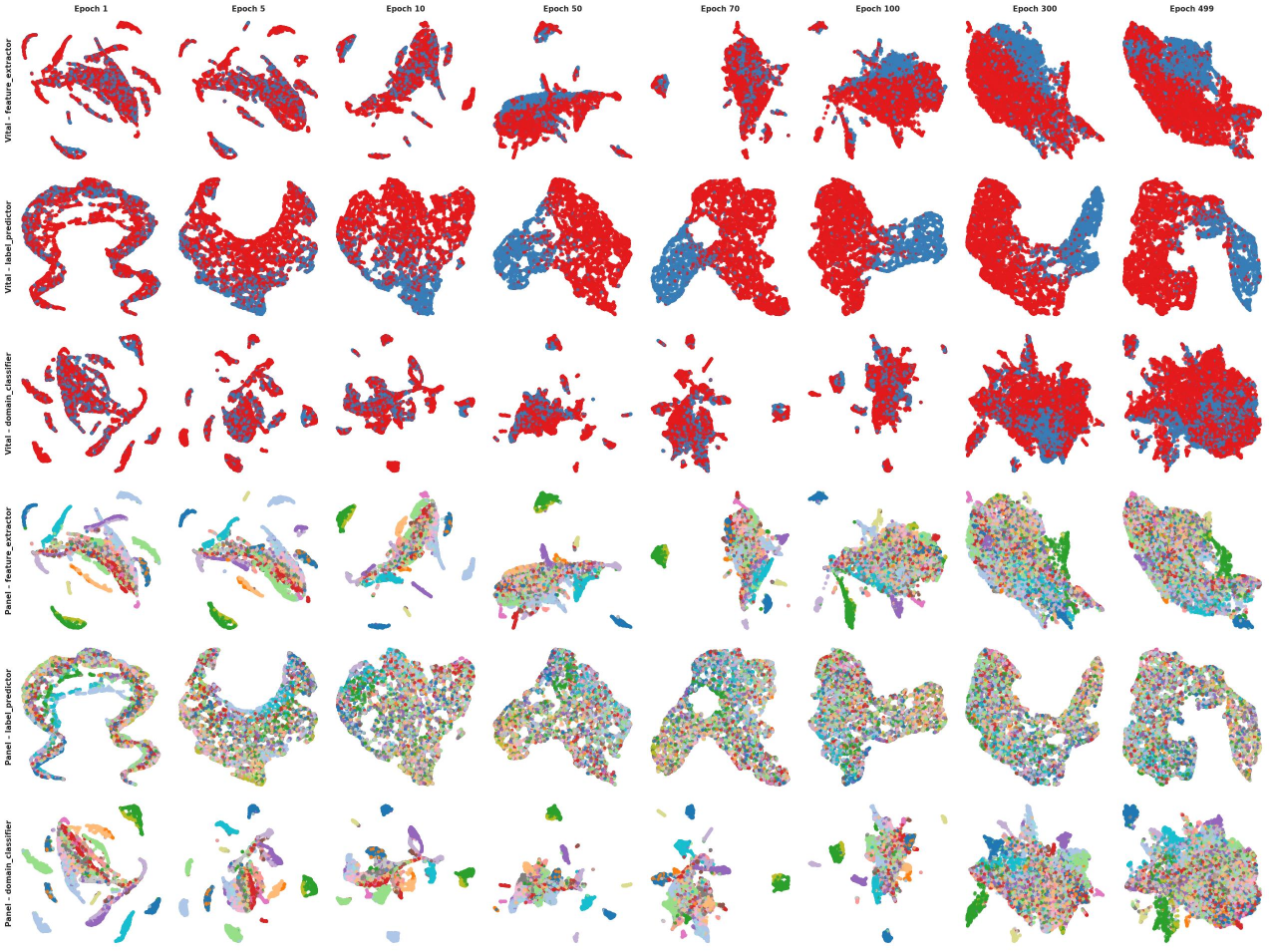}
    \caption{
    \textbf{2D UMAP projections of hidden activations reveal temporal evolution across DANN layers.} 2D UMAP visualizations of hidden layer activations from three distinct Domain-Adversarial Neural Network (DANN) layers across selected training epochs: 1, 5, 10, 50, 70, 100, 300, and 499. The top three rows display the UMAP projections colored by \textbf{vital status} (red = alive, blue = dead), while the bottom three rows show the same projections colored by \textbf{TCGA cancer type}. Each row corresponds to a different dropout layer in the DANN architecture: \texttt{feature\_extractor.dropout1}, \texttt{label\_predictor.dropout2}, and \texttt{domain\_classifier.dropout2}.
    }
    \label{fig:Figure4}
\end{figure}

When we colored the UMAP embeddings by vital status (the target variable), the label predictor layer consistently displayed the highest clustering scores over time, indicating that it learns highly discriminative features for classification. Conversely, when colored by the cancer panel (the domain variable), the same layer exhibited low clustering scores, suggesting that adversarial training successfully suppresses domain-specific signals. These trends confirm that the DANN framework encourages the network to prioritize class-relevant features while minimizing confounding effects from tissue origin.
To visualize these effects, Figure \ref{fig:Figure4} presents 2D UMAP embeddings of the same layers at different training epochs, colored by either vital status (top rows) or cancer panel (bottom rows). The progression of these plots illustrates that class separation strengthens over time in the label predictor layer, whereas domain separation progressively diminishes—especially in the domain classifier layer. This qualitative evidence supports the quantitative metrics, reinforcing that our DANN approach effectively learns domain-invariant representations.

\begin{figure}
    \centering
    \includegraphics[width=0.9\textwidth]{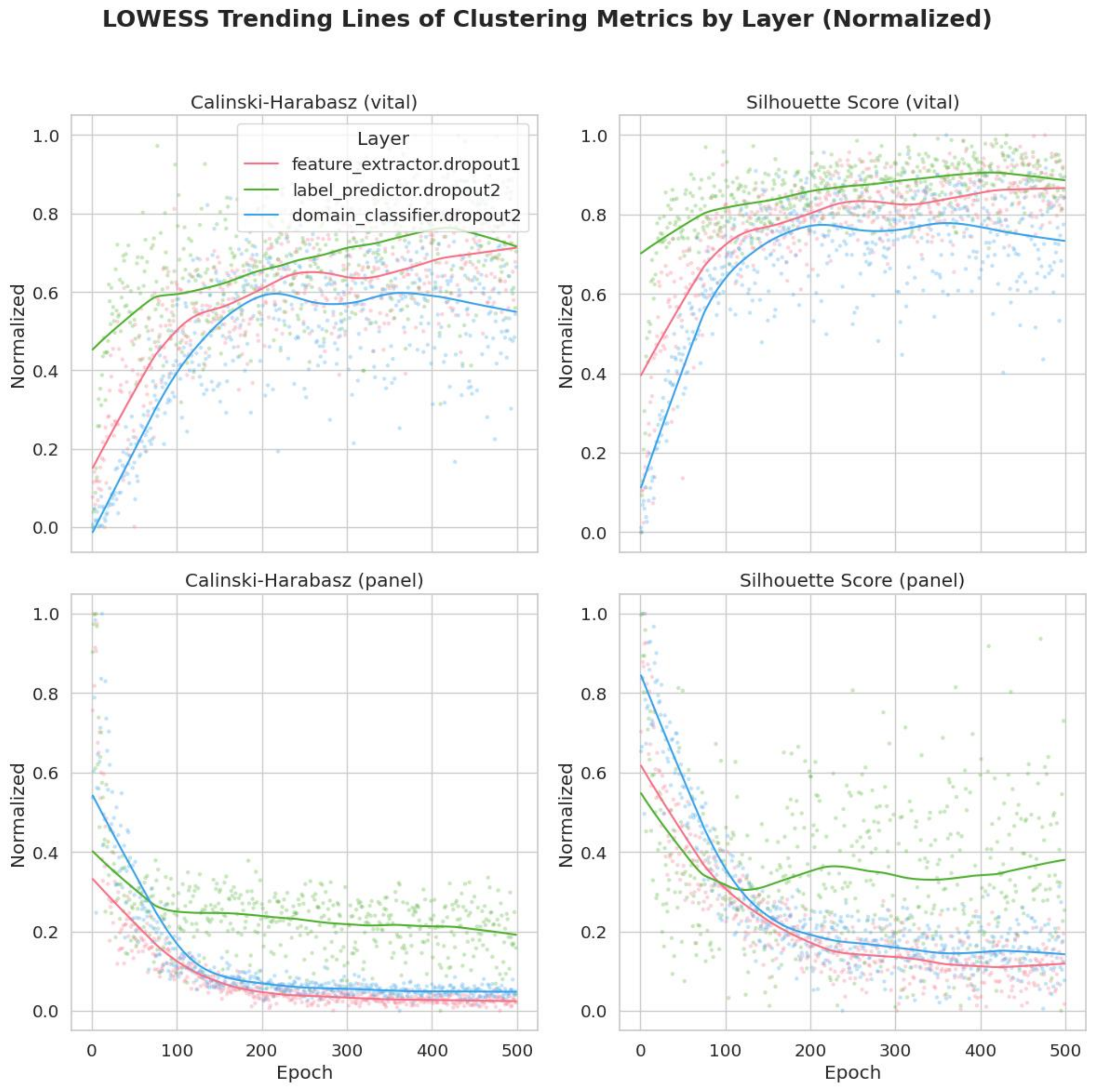}
    \caption{
    \textbf{Normalized clustering scores from SHAP-based UMAP projections reveal survival-relevant structure.} Normalized clustering scores (Calinski-Harabasz and Silhouette) computed on the 2D UMAP projections of SHAP values extracted from an XGBoost model trained on activations from different layers of the DANN architecture across training epochs. The top panels show clustering quality based on \textbf{vital status} (alive vs dead), while the bottom panels reflect clustering for \textbf{domain labels} (TCGA cancer types). SHAP values were computed for each layer separately and then projected using UMAP into two dimensions. Each line represents a different layer: \texttt{feature\_extractor.dropout1} (red), \texttt{label\_predictor.dropout2} (green), and \texttt{domain\_classifier.dropout2} (blue). By applying LOWESS smoothing, we observe a progressive increase in clustering structure related to survival outcomes, while clustering by cancer type diminishes, especially in the label predictor path. These trends reflect how interpretability patterns captured by SHAP align with the model’s learning dynamics, revealing disentanglement from tissue-of-origin signals and growing emphasis on features relevant to mortality. \textbf{This type of SHAP-based clustering evaluation could also be used as a criterion to determine an optimal stopping point during model training.}
    }
    \label{fig:Figure5}
\end{figure}

Beyond direct UMAP analysis (Figures \ref{fig:Figure3} and \ref{fig:Figure4}), we further assessed the representational utility of these layers by using them as inputs to an XGBoost classifier trained to predict vital status. We aimed to detect potential non-linear or combinatorial interactions in the neural network layer activations that UMAP might overlook. Since UMAP emphasizes preserving the dominant structure in the data, it may give more weight to high-magnitude activations, potentially obscuring low-magnitude but informative patterns. In contrast, SHAP values computed from an XGBoost classifier trained on these activations highlighted subtle but relevant features for mortality classification. Consequently, we computed SHAP (SHapley Additive exPlanations) values from the XGBoost model trained on each layer’s activations. Then, we used these values to gauge class-relevant structure via the same clustering metrics. As shown in Figure \ref{fig:Figure5}, the Calinski-Harabasz and Silhouette Scores derived from the SHAP-based UMAP embeddings converge more sharply and at earlier epochs, particularly for label\_predictor.dropout2 and feature\_extractor.dropout1. This behavior suggests that these layers encode features predictive of vital status sooner than what is discernible from raw UMAP projections. Moreover, domain-specific separation (cancer panel) is notably reduced in the SHAP-based space, confirming that these layers learn class-relevant yet domain-invariant features.

\begin{figure}
    \centering
    \includegraphics[width=0.9\textwidth]{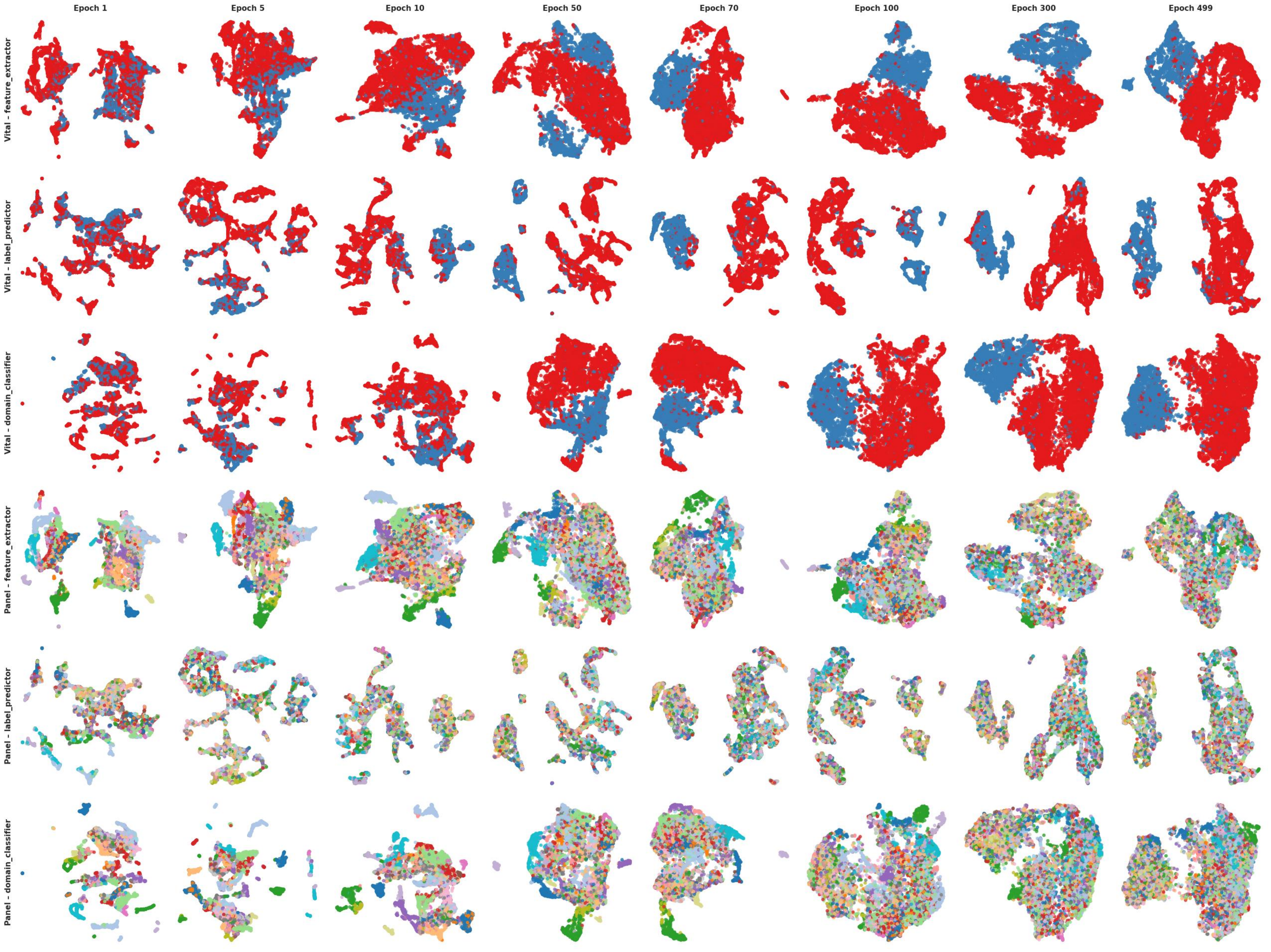}
    \caption{
    \textbf{Layer-specific SHAP-based UMAP visualizations reveal meaningful mortality-related patterns.}
    2D UMAP visualizations of SHAP values computed from an XGBoost classifier trained on activations from different layers of the Domain-Adversarial Neural Network (DANN). Each row corresponds to a specific dropout layer in the DANN architecture: \texttt{feature\_extractor.dropout1}, \texttt{label\_predictor.dropout2}, and \texttt{domain\_classifier.dropout2}. UMAP projections are shown for each layer, colored by \textbf{vital status} (red = alive, blue = dead; top panels) and by \textbf{TCGA cancer type} (bottom panels). Unlike raw activations, SHAP-based visualizations incorporate feature importance estimates, providing a more informative view of the decision-making process. This approach enhances the capacity to capture non-obvious relationships—including low-magnitude features—that may significantly influence the model's mortality classification performance, which is not easily discernible through unsupervised projections of activations alone.
    }
    \label{fig:Figure6}
\end{figure}

Complementing this, Figure \ref{fig:Figure6} presents the 2D UMAP visualizations derived from the SHAP values across training epochs. Notably, class separation emerges earlier and more distinctly than the direct activations shown in Figure \ref{fig:Figure4}. This critical observation underscores a fundamental distinction: High neuron activations do not inherently imply class relevance. A unit may exhibit strong activation without contributing meaningfully to the decision boundary, particularly in deep architectures where complex, non-linear interactions are at play.

By contrast, SHAP values capture the true contribution of each neuron to the model’s output, accounting for both direct effects and interaction terms. In this context, SHAP-based UMAPs act as functional projections—highlighting not just where activation occurs, but where importance lies. Therefore, the emergence of clearer class structure in SHAP embeddings marks the onset of representational convergence in the DANN architecture: the point at which internal representations begin to carry stable, class-relevant information as perceived by an external classifier (XGBoost).
Furthermore, this analysis highlights that conventional UMAP projections may underrepresent subtle but functionally significant patterns, primarily when those signals are distributed across neurons or dependent on interactions. The contrast between Figures \ref{fig:Figure4} and \ref{fig:Figure6} demonstrates the added interpretability and temporal resolution that explainability methods like SHAP offer when probing deep neural models. It emphasizes that model-informed attribution metrics can reveal the emergence of helpful structure well before it is visually apparent in raw activation space, offering a powerful complement to unsupervised manifold learning techniques.

\subsection{AI-Driven Feature Attribution: Limitations of Direct SHAP Analysis in Mortality Prediction} 
\label{subsec:AI-Driven}

\begin{figure}
    \centering
    \includegraphics[width=0.9\textwidth]{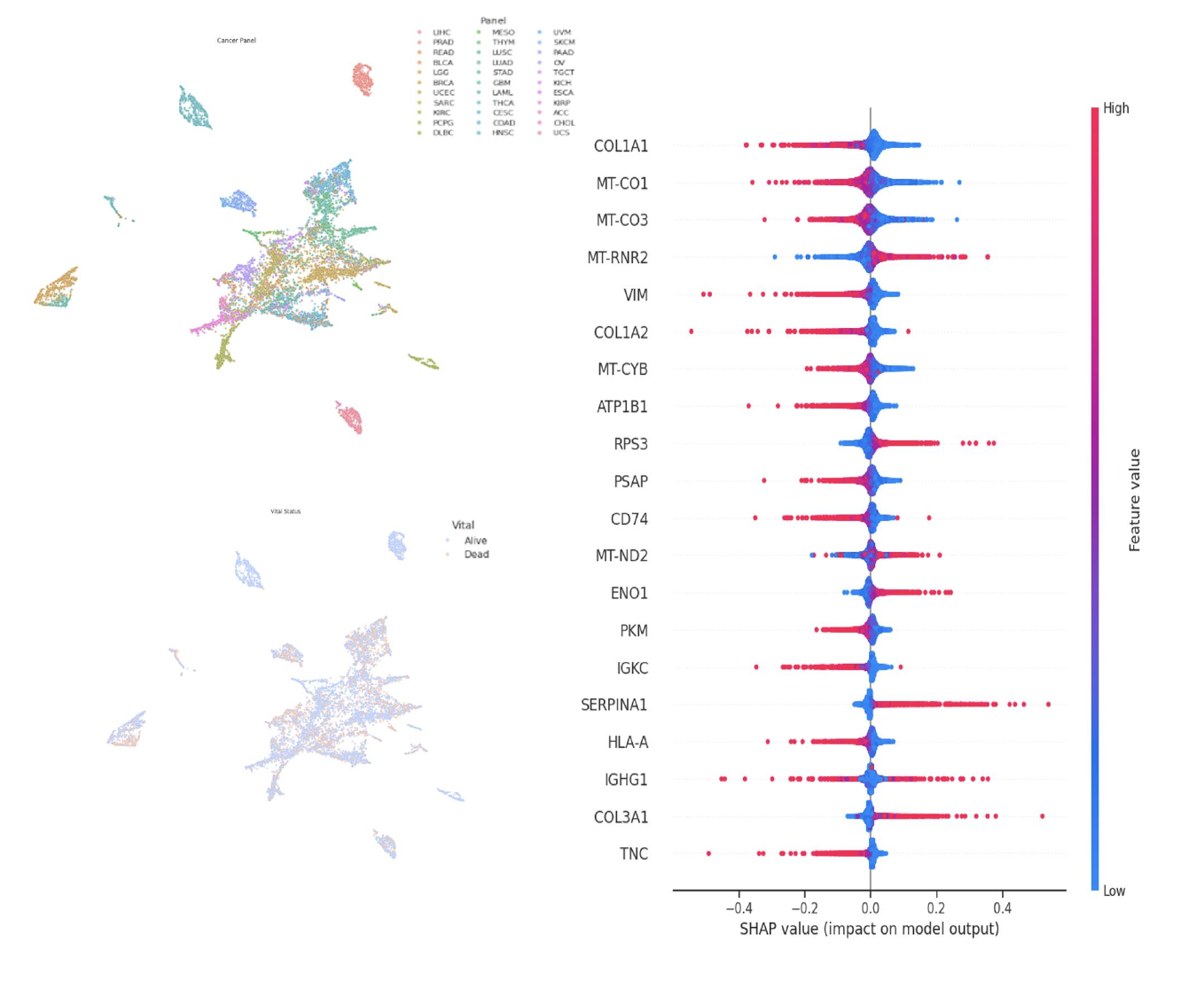}
    \caption{
    \textbf{Standard SHAP interpretation reveals persistent tissue-of-origin dominance.} 
    SHAP values computed on the output of the mortality classifier from the DANN model (vanilla SHAP) are shown. 
    Left: UMAP embedding of the samples colored by tissue of origin (top) and vital status (bottom) highlights that tissue-specific transcriptional signatures dominate the global structure. In contrast, survival status does not produce a clear separation. 
    Right: SHAP summary plot for top genes contributing to the mortality prediction shows that features related to tissue identity, such as mitochondrial and tumor microenvironment-associated genes, are highly influential. 
    These results indicate that, despite using a domain-adversarial network to mitigate tissue signal, vanilla SHAP remains confounded by tissue effects due to its reliance on raw input features.
    }
    \label{fig:Figure7}
\end{figure}

In Figure \ref{fig:Figure7}, we present the results of applying a vanilla SHAP analysis directly on the output of the mortality classifier in the DANN architecture. The UMAP visualizations of the SHAP values (left panels) reveal that the embedding space preserves a strong alignment with the tissue of origin (cancer panel), despite the model being adversarially trained to suppress domain-specific information. This preservation is evident from the clear cluster structure corresponding to distinct cancer types, as shown in the top panel. In contrast, the separation between vital status classes remains poorly defined in this representation (bottom panel), indicating that confounding tissue-specific signals heavily influence the SHAP values in this context.

The SHAP summary plot (right panel) further confirms this limitation: many of the top-ranked genes (e.g., COL1A1, MT-CO1, VIM, ATP1B1) are known to be tissue markers or highly expressed in specific tumor types rather than being consistent mortality predictors across cancers. These findings illustrate a critical pitfall of applying SHAP naively to the DANN classifier—while the model may have learned useful discriminative features, the direct attribution of output scores fails to disentangle class-relevant from domain-associated patterns, resulting in biased interpretations driven by residual domain signal.

We used another XAI method, namely Integrated Gradients \cite{sundararajan2017axiomatic}, but still observed domain dominance in the attributions (Supplementary \autoref{fig:FiguraSup1}). The failure of supervised clustering based on SHAP and Integrated Gradients to capture class-specific structure (vital status) may be primarily due to limitations inherent to the explanatory methods. While the DANN model reduces domain bias at the feature level, SHAP and IG remain agnostic to adversarial training and are applied directly to the final output. As a result, their attributions continue to reflect domain-associated patterns—such as tissue-specific gene expression—especially in settings where domain and class signals are entangled. This issue is exacerbated when explanation data used by SHAP are imbalanced, as shown by Liu et al.~\cite{liu2023balancedshap}, leading to biased attribution scores and misleading feature rankings.

\subsection{An Alternative Strategy: SHAP-Guided Clustering for Interpretable Latent Space Stratification}
\label{subsec:vanilla SHAP}

We trained an XGBoost classifier to predict vital status using the DANN feature extractor activations to evaluate the discriminative structure captured by our SHAP-guided clustering. SHAP values were then computed to generate an interpretable representation, which was projected into a low-dimensional manifold using UMAP and clustered with the Leiden algorithm. This process, applied to the full dataset, revealed five subpopulations potentially associated with distinct mortality-related mechanisms.

To assess the biological coherence of these clusters, a second XGBoost model was trained to predict Leiden membership using the original transcriptomic matrix. This model achieved a macro F1-score of approximately 0.6—a moderate but meaningful result indicating that the SHAP-based clusters capture structured and learnable biological variation (Supplementary Figure \ref{fig:FiguraSup2}). Since these clusters were derived from SHAP values computed on domain-invariant features, they reflect survival-related signals beyond tissue-of-origin effects (Figure~\ref{fig:Figure5}).

The emergence of well-separated Leiden clusters in the SHAP embedding suggests that DANN-extracted features retain biologically informative variation despite domain suppression. Clustering metrics (Calinski-Harabasz and Silhouette) improved more sharply and consistently across epochs when applied to SHAP embeddings compared to raw activations, particularly in layers closer to the label predictor (Figure~\ref{fig:Figure5}). Moreover, UMAP projections based on SHAP values yielded more apparent separation by survival status and diminished alignment with cancer type (Figure~\ref{fig:Figure6}), unlike projections from raw activations, which remained partially influenced by tissue structure.

\begin{figure}
	\centering
        \includegraphics[width=0.9\textwidth]{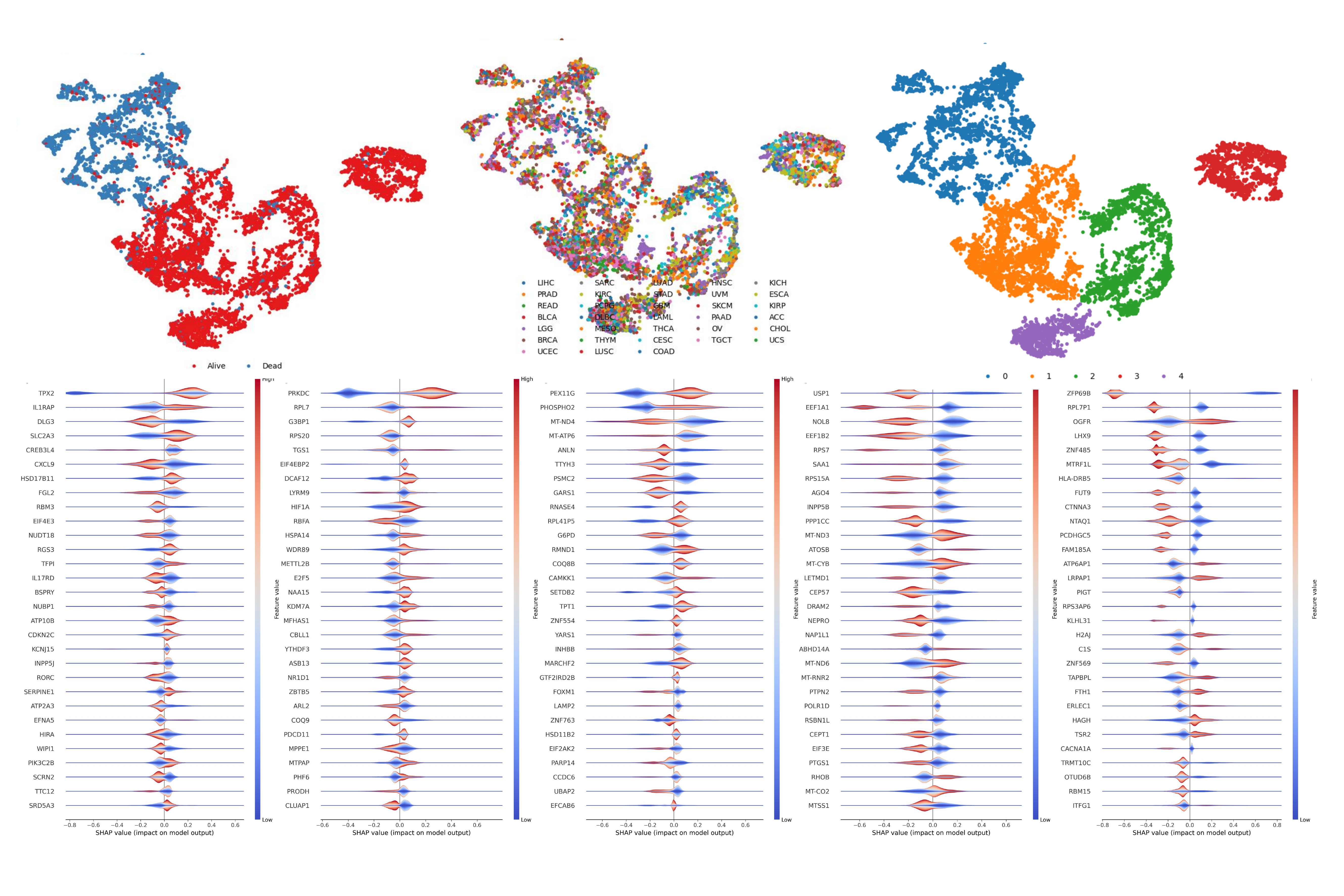}
	\caption{\textbf{Supervised Clustering of Feature Extractor Activations Using SHAP for Pan-Cancer Vital Status Subpopulation Characterization.} Supervised clustering of feature extractor activations using SHAP values derived from a multiclass classifier trained on Leiden subclusters (top right) based on activations from the DANN feature extractor. The UMAP projections at the top visualize the embedding space colored by vital status (left), cancer type (middle), and Leiden clusters (right). The SHAP summary plots (bottom) show the gene-level feature importance for each Leiden cluster (from cluster 0 to 4, left to right), indicating distinct molecular drivers associated with different mortality-related subpopulations across cancers. These results highlight the ability of the DANN-extracted representations to capture clinically meaningful stratification beyond tissue origin.}
	\label{fig:Figure8}
\end{figure}

We computed SHAP values from internal activations to address the limitations of direct SHAP interpretation on DANN outputs—still confounded by domain signals. This enabled a more biologically grounded view (Figure~\ref{fig:Figure7}) and enhanced resolution in detecting survival-associated subpopulations (Figure~\ref{fig:Figure8}). UMAP visualizations show that clusters derived from SHAP embeddings align more closely with survival status (left panel) than cancer type (middle panel), emphasizing the benefit of interpreting deep models at intermediate layers.

Gene-level SHAP summaries for each Leiden cluster uncovered distinct biological programs. We identified five gene clusters associated with tumor mortality across cancer types, each reflecting distinct biological processes independent of tissue origin.

Cluster 0 genes suggest a blend of immune-inflammatory and proliferative pathways. \textit{IL1RAP}, a mediator of IL-1 signaling, activates NF-$\kappa$B/AP-1, promoting inflammation and worse survival in cancers like pancreatic ductal adenocarcinoma \citep{cluster1_1}. \textit{CXCL9}, a T-cell-recruiting chemokine, correlates with CD8$^{+}$ T-cell infiltration and better outcomes in multiple tumors \citep{cluster1_2}. At the same time, \textit{FGL2}, an immunosuppressive factor, inhibits dendritic cells, fostering tumor progression and poor survival in glioblastoma and lung cancer \citep{cluster1_3}. \textit{TPX2}, a spindle assembly factor, drives proliferation and genomic instability, predicting shorter survival in solid tumors \citep{cluster1_4}. \textit{SLC2A3} (GLUT3), a glucose transporter, supports glycolytic metabolism and is a poor prognostic marker in colorectal and renal cancers \citep{cluster1_5}. \textit{RBM3}, a stress-response protein, is a favorable marker in some cancers, possibly countering aggressive traits here \citep{cluster1_6, cluster1_7, cluster1_8}.

Cluster 1 genes emphasize protein synthesis and stress resistance. Ribosomal proteins (\textit{RPS20}, \textit{RPL7}) and \textit{EIF4EBP2} indicate heightened translation, linked to tumor growth and metastasis in breast cancer \citep{cluster2_1}. \textit{EIF4EBP2}, a translational repressor, may reflect mTOR dysregulation, fueling oncogenic survival \citep{cluster2_2, cluster2_3}. \textit{PRKDC} (DNA-PKcs) enhances DNA repair, conferring resistance to genotoxic stress and worse outcomes in gastric cancer \citep{cluster2_4, cluster2_5}. \textit{G3BP1}, a stress granule component, promotes survival under stress, correlating with invasion and poor survival in hepatocellular carcinoma \citep{cluster2_6}. \textit{TGS1} and \textit{DCAF12} suggest RNA processing and protein turnover, supporting this aggressive phenotype \citep{cluster2_7, cluster2_8, cluster2_9}.

Cluster 2 genes focus on metabolism and proliferation. Mitochondrial genes \textit{MT-ND4} and \textit{MT-ATP6} indicate oxidative phosphorylation, linked to pancreatic cancer \citep{RN36, RN37, RN38}. \textit{PEX11G} is an unexplored gene in cancer but has been found to have a good prognosis in liver and renal cancer \citep{RN39}. \textit{ANLN}, a cytokinetic protein, is overexpressed in aggressive tumors like colorectal cancer, predicting worse prognosis \citep{RN40}. \textit{PSMC2}, a proteasome subunit, enhances protein turnover, with high expression tied to poor outcomes in gliomas \citep{RN41}. \textit{GARS1} expression was correlated with poor survival and had high diagnostic value in various tumor types \citep{RN42}. \textit{PHOSPHO2} and its read-through fusion with \textit{KLHL23} are implicated in cancer development, particularly in gastric cancer resistance to anticancer drug treatment \citep{RN43}.

Cluster 3 couples protein synthesis with stress and inflammation. Ribosomal genes (\textit{RPS7}, \textit{RPS15A}) and elongation factors (\textit{EEF1A1}, \textit{EEF1B2}) drive translation, with \textit{RPS7} promoting EMT in prostate cancer \citep{cluster2_1, RN31, RN32}. \textit{SAA1}, an inflammatory protein, correlates with an unfavorable overall survival in different cancers, especially in renal cell carcinoma and digestive cancer \citep{RN33}. \textit{USP1}, a DNA repair enzyme, enhances genomic stability, linked to chemoresistance in colorectal cancer \citep{RN34, RN35}. \textit{AGO4} and \textit{INPP5B} suggest epigenetic and signaling regulation.

Cluster 4 suggests immune engagement and differentiation. \textit{HLA-DRB5}, an MHC-II gene, indicates immunogenicity, correlating with better survival in breast cancer \citep{cluster4_1}. \textit{FUT9} drives stem-like traits and immune modulation in colon cancer \citep{cluster4_2}. \textit{CTNNA3}, a tumor suppressor, inhibits invasion in hepatocellular carcinoma \citep{cluster4_3}. \textit{OGFR} slows proliferation, with higher expression tied to less aggressive breast cancer \citep{cluster4_4, cluster4_5, cluster4_6}. \textit{LHX9} has been described as a tumor suppressor in some types of cancer and a potential oncogene in others. Its activity can influence cancer cell growth, proliferation, and metastasis by affecting various pathways, including glycolysis and the MAPK/PI3K pathways \citep{cluster4_7, cluster4_8, cluster4_9, cluster4_10, cluster4_11}. Inhibition of \textit{ZNF485} has been shown to promote the apoptosis of bladder cancer cells \citep{cluster4_12}. By clustering SHAP values rather than raw activations, we isolate latent functional dimensions most predictive of survival, reducing the influence of domain effects. This strategy enables an interpretable, stratified decomposition of the feature space into biologically distinct, survival-relevant subpopulations.

\vspace{-2em}

\begin{figure}
	\centering
        \includegraphics[width=0.9\textwidth]{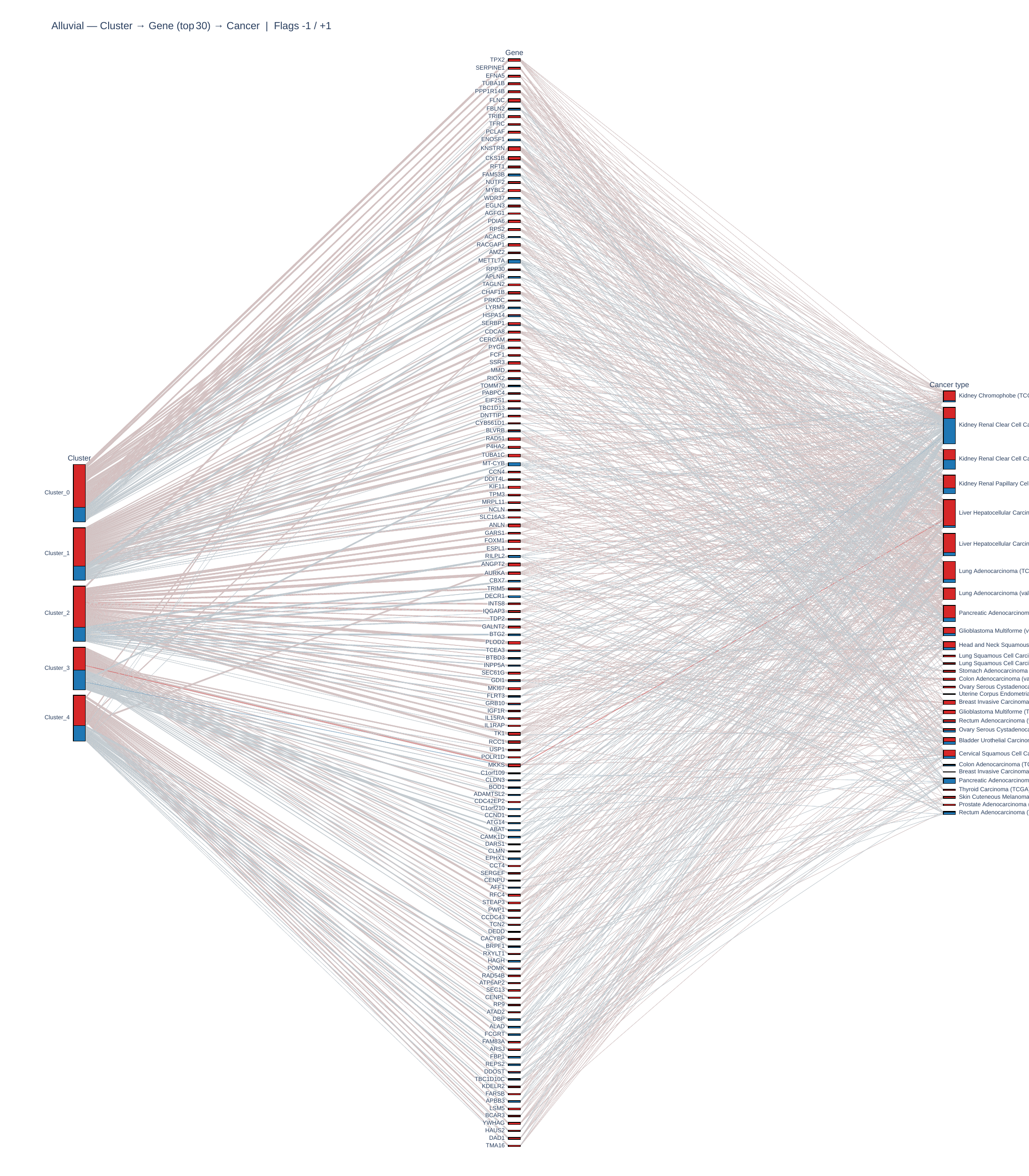}
	\caption{\textbf{Parallel‑categories diagram summarizing cluster‑specific prognostic genes across cancer types.} The alluvial plot depicts three categorical dimensions—gene‑expression cluster, gene symbol, and TCGA cancer type—to visualise how the top‑ranked genes from each cluster distribute across tumour cohorts and survival outcomes. Left axis (Clusters). Five transcriptomic clusters (Cluster 0–4) were obtained from our unsupervised analysis.  Middle axis (Genes). For every cluster, the 30 genes with the highest within‑cluster recurrence across cancers were retained (150 genes in total). Right axis (Cancer type). TCGA cohorts in which any of the displayed genes exhibits a prognostic association in the Human Protein Atlas (HPA v24.0) survival dataset.  Ribbons. Each ribbon connects a cluster to a gene and then to the cancer type(s) where that gene is prognostic. Ribbon width is proportional to the number of cancer cohorts represented, allowing quick visual assessment of gene ubiquity.  The prognostic direction is encoded red for unfavourable survival and blue for favourable survival.}
	\label{fig:Figure10}
\end{figure}
\vspace{0.5cm}

\section{Discussion}

In this work, we address \textbf{tissue-of-origin bias} in pan-cancer transcriptomics using a
\textbf{Domain-Adversarial Neural Network (DANN)} (see \S\ref{sec:Methods} and
Figures~\ref{fig:FiguraDANN}, \ref{fig:Figure2}). The central goal is to isolate survival-relevant
signals rather than spurious patterns driven by each tumor’s tissue of origin. We evaluated
multiple manifold representations, outlined below.

\subsection{ Integrative Interpretation: From Raw Activations to Layer-Aware Explanations}
\label{sec:IntegratedDiscussion}

\begin{itemize}
    \item As shown in Figures~\ref{fig:Figure3}--\ref{fig:Figure6}, we analyze how the DANN model evolves internal representations across training epochs, shifting from domain-driven to mortality-relevant structures. This transition is observed through both raw activations and SHAP-based projections derived from hidden layers.

    \item \textbf{Findings:}
    \begin{itemize}
        \item \textbf{Raw-activation UMAPs reveal domain-to-survival transition.} In the \textbf{raw-activation UMAPs} (Figure~\ref{fig:Figure4}), strong tissue-of-origin separation is observed in early epochs across all layers. Over time, the \textit{domain classifier} loses its ability to distinguish cancer types, while the \textit{label predictor} progressively separates patients by vital status. This dynamic reflects how adversarial training forces the shared feature extractor to eliminate tissue-specific signals, allowing survival-relevant features to become more salient.

        \item \textbf{Clustering metrics confirm representation shift.} Clustering metrics such as Calinski-Harabasz and Silhouette (\S\ref{subsec:Clustering}) corroborate this shift: in Figure~\ref{fig:Figure3}, the \emph{domain clustering} curve steadily declines—particularly in the label predictor—while \emph{vital status clustering} increases. This indicates a representational reorientation from lineage signals (i.e., which cancer type) toward survival signals (i.e., alive vs.\ dead), aligning with the DANN’s goal of domain-invariant learning.

        \item \textbf{SHAP-based manifolds highlight finer survival-related patterns.} In contrast, the \textbf{SHAP-based manifolds} ( Figures~\ref{fig:Figure5} and \ref{fig:Figure6}) exhibit a clearer, earlier emergence of mortality-relevant substructures. These patterns are less influenced by raw activation magnitude and more reflective of the model’s \emph{decision-making logic}. By directly attributing importance values to hidden-layer neurons, SHAP uncovers subtle but crucial survival signals that might be masked in the raw activations.

        \item \textbf{Low activation can still be high importance.} SHAP further reveals that features (or neurons) with low activation values can have disproportionately high importance scores, indicating that raw magnitude alone does not determine feature relevance. This is crucial for identifying mortality-associated signals that might be overlooked if one were guided solely by activation magnitude.
    \end{itemize}

    \item \textbf{Why It Matters:}
    \begin{itemize}
        \item \textbf{Validation of adversarial training for pan-cancer analysis.} These observations validate that adversarial training achieves the goal of creating latent spaces free from tissue signals, while still preserving information relevant to survival. From a biological standpoint, this implies that the DANN is capturing molecular patterns shared across different cancer types, which is essential for uncovering \textbf{pan-cancer mortality mechanisms}.

        \item \textbf{SHAP-based UMAPs enhance interpretability.} SHAP-based UMAPs offer a richer, more explanatory view of the model’s internal logic. This is particularly relevant when searching for \textbf{pan-cancer biomarkers}, as it reveals subpopulations with similar clinical outcomes that may not cluster by tissue of origin, but rather by shared molecular signatures.

        \item \textbf{Early stopping and overfitting prevention.} The saturation of SHAP-based clustering metrics (Figure~\ref{fig:Figure5}) suggests that these trends could serve as a reliable criterion for \textbf{early stopping}, preventing the model from overfitting noise in the data. Maximizing the “purity” of survival representations helps improve the classifier’s generalization capacity.

        \item \textbf{Combining raw activations and SHAP for a comprehensive view.} While raw activations reveal the gradual suppression of tissue signals, overlaying them with SHAP provides \textbf{an integral understanding} of which neuronal elements truly “matter” for mortality. This allows a more detailed examination of how the model reconciles conflicting signals, enhancing both interpretability and robustness.
    \end{itemize}
\end{itemize}

\subsection{Standard (Vanilla) SHAP at the Model Output}
\label{sec:standardshap}

\begin{itemize}
    \item As described in Section~\ref{subsec:vanilla SHAP} and illustrated in Figure~\ref{fig:Figure7}, we applied SHAP \emph{directly} to the final output of the DANN model, using the \textbf{raw input gene expression matrix} as the explanatory feature space. This strategy—commonly referred to as \emph{vanilla SHAP}—aims to explain the model's mortality predictions by estimating each gene's contribution. However, the SHAP-derived manifold revealed persistent \textbf{tissue-of-origin dominance}, with samples clustering primarily by cancer type rather than survival status. Additionally, \textbf{Integrated Gradients (IG)} showed similar patterns of cancer-type-driven separation.

    \item \textbf{Findings:}
    \begin{itemize}
        \item Strong tissue signals remained evident in the explainable manifolds generated by both vanilla SHAP and Integrated Gradients.
        
        \item This effect was further amplified by class imbalance, a well-documented limitation of SHAP, which tends to bias feature attributions toward overrepresented classes—in this case, cancer types with larger sample sizes \citep{liu2023balancedshap}. This issue is particularly problematic because both tissue type and vital status classes are imbalanced. As a result, SHAP values are skewed by the background distribution, making it difficult to generate fair or representative explanations without synthetic balancing strategies—an approach not pursued here due to the biological complexity of the data.
    
        \item Integrated Gradients (IG), a path-based interpretability method that computes attributions by integrating gradients from a baseline to the actual input, also failed to isolate survival-specific signals. Despite its theoretical robustness for deep neural networks, IG likewise revealed manifolds dominated by tissue-of-origin patterns when visualized using dimensionality reduction techniques (e.g., UMAP).
    
        \item Both SHAP and IG produced \textbf{convergent global attributions} (top importance features), reinforcing the conclusion that input-level explanation methods remain confounded by strong domain signals, and fail to capture survival-relevant features effectively.
    \end{itemize}

    \item \textbf{Why It Matters:}
    \begin{itemize}
        \item These results demonstrate that applying interpretability methods directly to the raw input features—even after adversarial training—can produce \textbf{misleading or biased explanations}, as they continue to reflect confounding domain signals that the DANN was trained to suppress.
        \item The \textbf{vanilla SHAP-derived manifold} (Figure~\ref{fig:Figure7}) exhibited clearer separation by tissue than by survival outcome, emphasizing how input-level representations remain entangled with domain-specific variance.
        \item This finding underscores the need for \textbf{layer-aware interpretability}, which focuses on internal network activations rather than input-level attributions. Hidden layers encode progressively refined and potentially domain-invariant representations that are better suited for understanding the model’s actual decision process.
        \item Interrogating internal representations—particularly those from the feature extractor and label predictor—allows us to assess whether the DANN has successfully extracted survival-relevant information while mitigating tissue bias.
    \end{itemize}
\end{itemize}

\subsection*{Overall Takeaways}

\begin{itemize}
    \item \textbf{Raw-activation UMAPs} (no SHAP) confirm domain-signal suppression but may hide subtle mortality-related features if overshadowed by high-magnitude units.

    \item \textbf{Layer-aware SHAP Manifolds} (SHAP on hidden activations) offer a clearer view of survival-specific signals unburdened by tissue-of-origin bias, thus revealing meaningful subpopulations with shared molecular traits linked to mortality.

    \item \textbf{Vanilla SHAP on raw inputs} is still prone to tissue confounding, underscoring that we must move beyond naive input-level explanations in adversarially trained models.
\end{itemize}

\subsection{Other Related Work}

Understanding the internal dynamics of neural networks is critical for interpreting how deep models learn and generalize, particularly in complex biomedical tasks like pan-cancer survival analysis. Recent studies have explored visualization and explainability techniques to probe these dynamics, but few address the challenges of high-dimensional, heterogeneous domains such as transcriptomic data across multiple cancer types. This section reviews relevant work and highlights how our study advances domain-adversarial neural networks (DANNs) and interpretability in this context.

A recent study by \cite{xu2025explainable} evaluated DANNs using t-SNE on the last hidden layer to visualize adaptation between two financial domains, based on low-dimensional tabular data with ~30 features. They showed reduced domain separability post-adaptation but were limited to a simplified scenario with two domains and low input dimensionality. In contrast, our work tackles domain adaptation across 33 cancer types from the TCGA pan-cancer dataset, using high-dimensional transcriptomic profiles with over 39,000 biotype transcripts. We extend explainability beyond input features by applying SHAP to internal hidden layer activations, offering a richer, structurally informed view of how domain-invariant representations emerge and capture survival-relevant signals beyond tissue-specific biases.

Visualization techniques like \textit{Multislice PHATE}, introduced by \cite{gigante}, reveal how neural network representations evolve from disorganized to semantically meaningful structures over training epochs. Inspired by this, we analyze how feature representations in DANNs evolve across tasks—domain discrimination and target classification—during training. This perspective enables us to trace the internal learning dynamics of domain-adversarial models, shedding light on their behavior beyond final predictions.

Initially, we considered applying SHAP values to internal neurons for a target output to be novel. However, \cite{Fidel} previously used SHAP on the penultimate layer of a pre-trained classifier (e.g., CIFAR-10) to create \textit{XAI signatures}—flattened matrices encoding each neuron’s contribution to each class. These signatures trained a binary neural network to detect adversarial examples. Unlike \cite{Fidel}, we apply SHAP directly to multiple internal layers of a DANN to evaluate how they encode domain-specific (e.g., tissue-of-origin) and task-specific (e.g., survival outcome) signals. This approach treats SHAP as an investigative tool to assess the disentanglement of competing information sources, providing layer-wise interpretability into how adversarial learning shapes internal representations. To our knowledge, this is the first work to systematically analyze a DANN’s internal feature space using SHAP in this way.

This layer-wise interpretability is particularly valuable in pan-cancer modeling, where tissue-of-origin signals often dominate transcriptomic profiles, confounding the learning of outcome-relevant features. By applying SHAP across multiple layers, we track how domain and task information propagate, ensuring clinically meaningful signals—like those tied to patient survival—are preserved and prioritized. This framework enhances the accuracy of mortality classification and the reliability of survival time regression models across diverse tumor types.

In conclusion, for robust \textbf{pan-cancer survival analysis}, we recommend combining \textbf{DANNs} with \textbf{layer-level interpretability} (\S\ref{subsec:Clustering}) to isolate mortality-related genes from tissue-driven noise. This pipeline effectively captures \textbf{universal survival biomarkers} across cancer types, demonstrating the power of manifold-based explainability to uncover fine-grained signals hidden by strong tissue-of-origin biases.

\section{Methods}
\label{sec:Methods}
\subsection{Data TCGA}
Data were obtained from https://github.com/resendislab/proliferation, which directs users to an R dataset. This dataset contains The Cancer Genome Atlas (TCGA) Gene Expression data, where both clinical and raw matrices can be found in an R database named "tcga.rds." The following lines of R code were used to extract clinical information (including tumor status and vital status, among others) as well as the RNA-seq matrix, which encompasses 11,093 samples and 60483 GeneIDs (Stable IDs are created in the form of ENS using the following R code):
```\\
tcga <- readRDS("tcga.rds"),  tcga\$rnaseq\$counts and (tcga[["clinical"]] \\
```
Next, in Python 3, we transposed the data using the following command: `pd.read\_csv('raw\_T.csv')`. We converted 60483 columns (ENS and GeneID) into 41124 gene symbols. Then, we collapsed the data by summing the gene expression values of duplicate gene symbols, resulting in a final count of 39979 gene symbols. Given the capabilities of XGBoost modeling to handle very high-dimensional data and its robustness to scale magnitudes, we decided to use all gene dimensions. This includes various biotypes, such as protein-coding genes, lncRNAs, pseudogenes, and miRNAs, among others.

All datasets, preprocessing scripts, and model implementations used in this study are publicly available at the following GitHub repository: \url{https://github.com/resendislab/Pancancer_project}.

\subsection{TCGA UMAP cancer type dominance.}
To visualize the dominant structure of cancer types in transcriptomic data, we applied dimensionality reduction to the raw gene expression matrix using a standard pipeline. First, the expression matrix was log-transformed to stabilize variance. We then performed Principal Component Analysis (PCA) to reduce the data to 50 dimensions, capturing the major axes of transcriptomic variation. Subsequently, we applied Uniform Manifold Approximation and Projection (UMAP) to the PCA output, reducing it to two dimensions for visualization. UMAP was configured with 30 neighbors and a minimum distance of 0.3. This visualization, shown in Figure \ref{fig:Figure1}, was used solely to illustrate the strong clustering of samples by tissue of origin (i.e., TCGA cancer types), highlighting the dominance of cancer-type-specific expression patterns in unsupervised low-dimensional embeddings

\subsection{Domain Adversarial Neuronal Network Architecture theory  theory}

\begin{figure}
	\centering
    \includegraphics[width=0.9\textwidth]{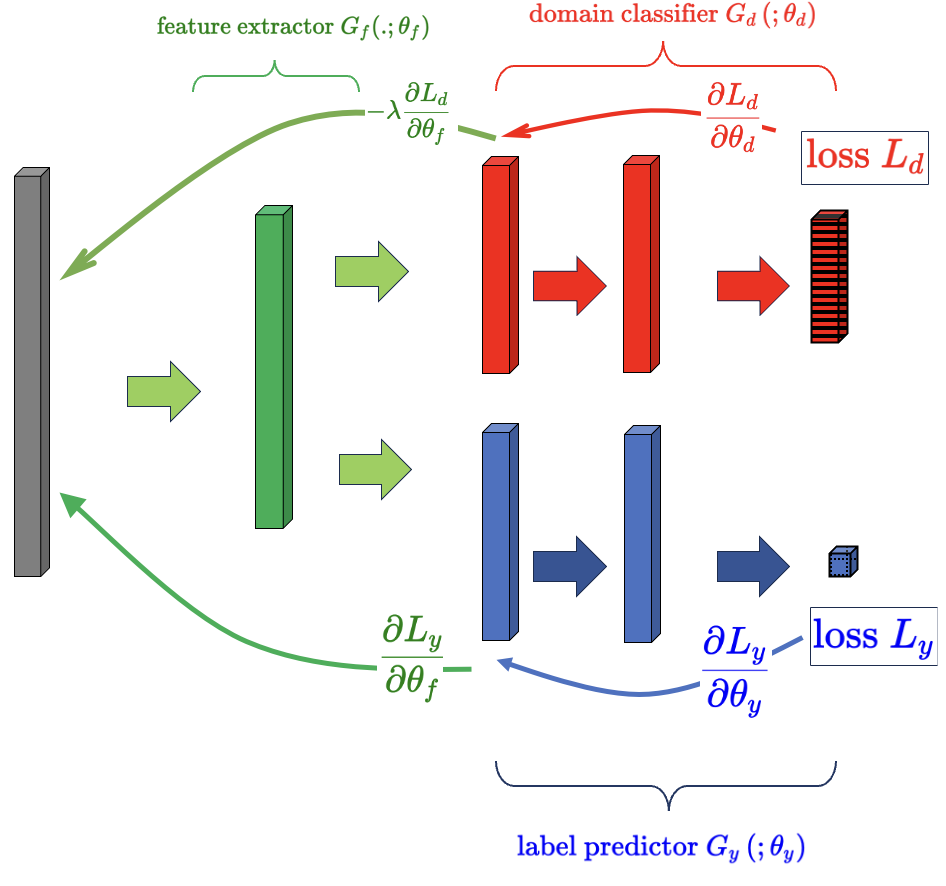}
	\caption{Architecture of the Domain-Adversarial Neural Network (DANN). The figure illustrates the architecture of a Domain-Adversarial Neural Network (DANN), comprising three main components: the feature extractor (green), the label predictor (blue), and the domain classifier (red). The feature extractor maps input data into a latent representation, which is then used by both the label predictor and the domain classifier. The label predictor is trained to minimize the classification loss by optimizing its parameters using gradients. Simultaneously, the domain classifier attempts to distinguish between different domains, minimizing its loss with respect to the calculated value. To enforce domain invariance, a gradient reversal layer (GRL) applies a negative scaling factor to the domain classification loss gradient, ensuring that the feature extractor learns domain-invariant representations by modifying the gradient. This adversarial process enables the model to generalize across domains while maintaining high classification accuracy. The architecture is commonly used in domain adaptation tasks, such as survival prediction across multiple cancer datasets.}
	\label{fig:FiguraDANN}
\end{figure}

To address the domain adaptation problem in deep learning, we implemented a Domain-Adversarial Neural Network (DANN) designed to learn domain-invariant representations through adversarial training (Figure \ref{fig:FiguraDANN}). The model architecture consists of three main components:

\begin{enumerate}
    \item \textbf{Feature Extractor} $G_f(x; \theta_f)$: A deep learning model parameterized by $\theta_f$ that learns a shared representation for both source and target domains. It is responsible for transforming raw input data into a feature space that is useful for classification while attempting to make the features indistinguishable between domains.

    \item \textbf{Label Predictor} $G_y(G_f(x); \theta_y)$: A network that assigns class labels to input data, with parameters $\theta_y$. This component is trained using labeled data from the source domain and optimizes a classification loss.

    \item \textbf{Domain Classifier} $G_d(G_f(x); \theta_d)$: A discriminator that attempts to distinguish whether a sample originates from the source or target domain, with parameters $\theta_d$. This classifier helps to ensure that the extracted features are domain-invariant by applying an adversarial loss.
\end{enumerate}

\subsection{Loss Function}

The network is trained by optimizing the following objective function, which balances the main task and domain adaptation:

\[
E(\theta_f, \theta_y, \theta_d) = \frac{1}{n} \sum_{i=1}^{n} \mathcal{L}_y^i (\theta_f, \theta_y) - \lambda \left( \frac{1}{n} \sum_{i=1}^{n} \mathcal{L}_d^i (\theta_f, \theta_d) + \frac{1}{n'} \sum_{i=n+1}^{N} \mathcal{L}_d^i (\theta_f, \theta_d) \right)
\]

where:

\begin{itemize}
    \item $\mathcal{L}_y^i(\theta_f, \theta_y) = L_y(G_y(G_f(x_i; \theta_f); \theta_y), y_i)$ is the classification loss in the source domain.
    \item $\mathcal{L}_d^i(\theta_f, \theta_d) = L_d(G_d(G_f(x_i; \theta_f); \theta_d), d_i)$ is the domain classifier loss, where $d_i = 0$ for source samples and $d_i = 1$ for target samples.
    \item $\lambda$ is a fixed hyperparameter that controls the influence of adversarial regularization.
\end{itemize}

\subsection{Adversarial Optimization}

DANN is trained using a minimax optimization framework, aiming to find the saddle point by solving:

\[
(\hat{\theta_f}, \hat{\theta_y}) = \arg \min_{\theta_f, \theta_y} E(\theta_f, \theta_y, \hat{\theta_d})
\]
\[
\hat{\theta_d} = \arg \max_{\theta_d} E(\hat{\theta_f}, \hat{\theta_y}, \theta_d)
\]

This means that while the feature extractor and label predictor minimize classification loss, the domain classifier maximizes the discrepancy between the source and target distributions. The feature extractor is simultaneously trained to fool the domain classifier by producing domain-invariant representations.

Although the original DANN framework applies \textbf{stochastic gradient descent (SGD)}, \textbf{our implementation diverges from this approach} by utilizing \textbf{AdamW} instead of SGD for all model components. The AdamW update equations are:

\[
m_t = \beta_1 m_{t-1} + (1 - \beta_1) g_t
\]
\[
v_t = \beta_2 v_{t-1} + (1 - \beta_2) g_t^2
\]
\[
\hat{m}_t = \frac{m_t}{1 - \beta_1^t}, \quad \hat{v}_t = \frac{v_t}{1 - \beta_2^t}
\]
\[
\theta \leftarrow \theta - \eta \left( \frac{\hat{m}_t}{\sqrt{\hat{v}_t} + \epsilon} + \lambda \theta \right)
\]

where $\eta$ is the learning rate, $\beta_1$ and $\beta_2$ are momentum terms. In this context, $\lambda$ represents the weight decay coefficient and should not be confused with $-\lambda$, which is a separate hyperparameter that controls the influence of adversarial regularization.

\subsection{Gradient Reversal Layer (GRL) and Optimizer Choice}

To enforce adversarial optimization, a \textbf{Gradient Reversal Layer (GRL)} is applied to the feature extractor. The GRL acts as a regular neural network layer during the forward pass:

\[
R(x) = x
\]

However, during backpropagation, it negates the gradient and scales it by $-\lambda$, forcing the feature extractor to learn domain-invariant representations:

\[
\frac{\partial \mathcal{L}_d}{\partial \theta_f} \leftarrow -\lambda \frac{\partial \mathcal{L}_d}{\partial \theta_f}
\]

Using AdamW instead of SGD (stochastic gradient descent)introduces adaptive learning rates, which may affect the learning dynamics of the feature extractor and domain classifier differently compared to standard SGD-based training. The adaptive moment estimation of AdamW helps stabilize training while preserving the adversarial nature of GRL.

\subsection{Implementation and Hyperparameter Configuration}

The model was implemented using \textbf{PyTorch}, with a deep neural network architecture consisting of fully connected layers, ReLU activation functions, and dropout regularization. The \textbf{AdamW} optimizer was used with a learning rate of $10^{-4}$, and the value of $\lambda$ was fixed throughout the training process. Dropout was applied to improve generalization and prevent overfitting.

For training, we used mini-batch stochastic gradient descent, and the domain classifier was updated at the same rate as the feature extractor to ensure consistent adversarial learning. Batch normalization was also applied to stabilize training and improve convergence.

\subsection{DANN Architecture Components}

The Domain-Adversarial Neural Network (DANN) consists of three main components: the Feature Extractor ($G_f$), the Label Predictor ($G_y$), and the Domain Classifier ($G_d$). The architecture is designed to learn domain-invariant feature representations while simultaneously optimizing for the classification of vital status and domain (panel) prediction. Below, we describe the internal layers and the operations applied to each component.

\subsubsection*{Feature Extractor ($G_f$)}

\begin{itemize}
    \item fc1: Fully connected layer (input $\rightarrow$ 1000)
    \item bn1: Batch normalization
    \item dropout1: Dropout (p=0.1)
    \item Leaky ReLU activation
\end{itemize}

\subsubsection*{Label Predictor ($G_y$)}

\begin{itemize}
    \item fc1: Fully connected (1000 $\rightarrow$ 1000), followed by bn1, dropout1, Leaky ReLU
    \item fc2: Fully connected (1000 $\rightarrow$ 1000), followed by bn2, dropout2, Leaky ReLU
    \item fc3: Fully connected (1000 $\rightarrow$ 1), followed by sigmoid activation
\end{itemize}

\subsubsection*{Domain Classifier ($G_d$)}

\begin{itemize}
    \item fc1: Fully connected (1000 $\rightarrow$ 1000), followed by bn1, dropout1, Leaky ReLU
    \item fc2: Fully connected (1000 $\rightarrow$ 1000), followed by bn2, dropout2, Leaky ReLU
    \item fc3: Fully connected (1000 $\rightarrow$ \#domains), followed by softmax activation
\end{itemize}

\subsubsection*{Reverse Gradient Layer and Optimization}

A Reverse Gradient Layer (GRL) is placed between $G_f$ and $G_d$ to apply adversarial training by multiplying the gradients by $-\lambda$ during backpropagation.

The model uses:
\begin{itemize}
    \item BCE loss for $G_y$ (binary classification)
    \item CE loss for $G_d$ (multi-class classification)
    \item AdamW optimizer with:
        \begin{itemize}
            \item Learning rate: 0.01
            \item Betas: (0.9, 0.99)
            \item Weight decay: 0.01
        \end{itemize}
\end{itemize}
\begin{table}[h!]
\small
\centering
\caption{Summary of the DANN architecture components, layers, neurons, and operations.}
\renewcommand{\arraystretch}{1.2}
\begin{tabular}{|p{3.5cm}|p{3.5cm}|p{1.5cm}|p{6cm}|}
\hline
\textbf{Component} & \textbf{Layer} & \textbf{Neurons} & \textbf{Operation} \\
\hline
Feature Extractor ($G_f$) & Fully Connected (fc1) & 1000 & Projects input to 1000-D space \\
                          & Batch Norm (bn1) & -- & Stabilizes activations \\
                          & Dropout (dropout1) & -- & 10\% dropout \\
                          & Leaky ReLU & -- & Allows small negative gradients \\
\hline
Label Predictor ($G_y$)   & Fully Connected (fc1) & 1000 & Initial classifier layer \\
                          & Batch Norm (bn1) & -- & Stabilizes learning \\
                          & Dropout (dropout1) & -- & 10\% regularization \\
                          & Fully Connected (fc2) & 1000 & Refines features \\
                          & Batch Norm (bn2) & -- & Enhances training \\
                          & Dropout (dropout2) & -- & Prevents overfitting \\
                          & Fully Connected (fc3) & 1 & Survival output \\
                          & Sigmoid & -- & Converts to probability \\
\hline
Domain Classifier ($G_d$) & Fully Connected (fc1) & 1000 & Maps to domain-relevant space \\
                          & Batch Norm (bn1) & -- & Improves convergence \\
                          & Dropout (dropout1) & -- & 10\% regularization \\
                          & Fully Connected (fc2) & 1000 & Refines features \\
                          & Batch Norm (bn2) & -- & Improves stability \\
                          & Dropout (dropout2) & -- & Prevents overfitting \\
                          & Fully Connected (fc3) & \# domains & Domain logits \\
                          & Softmax & -- & Probability over domains \\
\hline
GRL & Gradient Reversal ($\lambda$) & -- & Reverses gradient to enforce domain invariance \\
\hline
\end{tabular}
\end{table}

\subsection{Evaluation of DANN architecture }

To evaluate the DANN architecture, the loss and precision of domain and vital status classifications were calculated for each epoch up to 499 epochs. Additionally, 5-fold cross-validation was applied in each epoch, and the results were visualized for both the domain and vital status classifiers.

DANN performance evaluations vital status prediction and pan-cancer type classification.
To evaluate the performance and generalizability of the Domain-Adversarial Neural Network (DANN), we applied a 5-fold stratified cross-validation strategy. This approach ensured that each fold preserved the original proportion of samples per class (vital status), allowing a balanced and robust evaluation. For each fold, the model was trained from scratch using a training subset (80\%) and evaluated on the remaining validation subset (20\%), rotating folds across iterations.

The improved Domain-Adversarial Neural Network (DANN) was trained using transcriptomic data with a batch size of 128 over 499 epochs. The architecture consisted of a feature extractor, a label predictor, and a domain classifier, each with two hidden layers of 1000 units, batch normalization, and dropout regularization (rate = 0.1). He initialization was applied to all linear layers. The model was optimized using AdamW with a learning rate of 0.01, $\beta_1 = 0.9$, $\beta_2 = 0.99$, weight decay of 0.01, and $\varepsilon = 1 \times 10^{-8}$, without AMSGrad. A constant adversarial coefficient of $\lambda = 0.01$ was used during training, controlling the gradient reversal strength to enforce domain invariance. Binary cross-entropy was used for the label prediction loss, while cross-entropy was used for domain classification. Training and validation sets were stratified by vital status, and the class labels for domain and survival outcome were encoded using label encoders.

We monitored four key metrics during training and validation for both tasks:

\begin{itemize}
    \item Label classification loss (BCELoss) and accuracy (based on a 0.5 threshold on sigmoid outputs)
    \item Domain classification loss (CrossEntropyLoss) and accuracy (based on the predicted cancer panel).
\end{itemize}

Cross-validation results were visualized by plotting training/validation losses and accuracies for both label and domain classifiers across all folds. This evaluation confirmed the model's consistency across data partitions and supported the effectiveness of the adversarial framework in learning survival-relevant, domain-invariant representations.

Finally, we trained again using 80\% of the training data and saved the model at the last batch of each epoch to enable detailed tracking of training dynamics over time. High validation label accuracy reflects the model’s ability to predict survival status effectively, while a progressive decrease in domain accuracy suggests that the feature extractor is successfully suppressing tissue-specific signals—one of the key objectives of domain-adversarial training.

\subsection{DANN clustering metrics evaluations.}
To evaluate the evolution of clustering structure in the learned representations during training, we computed the Calinski-Harabasz and Silhouette Scores across 499 epochs. These metrics were applied to two types of 2D UMAP projections: (1) UMAP embeddings derived from the activations of the last transformation layer (dropout) in each DANN branch—namely feature\_extractor.dropout1, label\_predictor.dropout2, and domain\_classifier.dropout2; and (2) UMAP embeddings computed from SHAP values obtained via an XGBoost classifier trained to predict vital status using the corresponding layer’s activations as input. 

All score values were normalized using Min-Max scaling to allow fair comparison across layers and metrics. To visualize temporal trends, we applied LOWESS (Locally Weighted Scatterplot Smoothing) with a smoothing parameter frac=0.3, which controls the fraction of data used for each local regression and provides a balance between capturing local variation and ensuring smooth global trends. This procedure enabled systematic tracking of class- and domain-related structure in both raw and SHAP-transformed representations throughout DANN training.

\subsection{Feature Attribution from Transcriptome Matrix Input Using SHAP and Integrated Gradients.}

To evaluate feature-level contributions to the prediction of vital status, we applied model explainability techniques using the original high-dimensional transcriptomic input and the output of the DANN label predictor. Specifically, we used two attribution methods: (1) vanilla SHAP values, computed directly from the model output using the original expression matrix, and (2) Integrated Gradients (IG), implemented via Captum (from captum.attr import IntegratedGradients), which integrates gradients along a path from a baseline to the input. These attributions were used exclusively for post hoc model interpretation to examine the distribution of input feature relevance. To assess whether the resulting attribution patterns remained influenced by cancer type, we reduced the dimensionality of both the SHAP and IG values using UMAP and visualized the resulting embeddings colored by cancer panel. This step allowed us to evaluate whether the model’s explanations were still confounded by tissue-of-origin signals, despite being adversarially trained to suppress them. 

\subsection{SHAP-Guided Clustering for Latent Space Stratification}
For unsupervised identification of survival-relevant subpopulations, we applied Leiden clustering on a UMAP projection derived from SHAP values. SHAP values were computed from an XGBoost classifier trained to predict vital status using the activations of the feature\_extractor.dropout1 layer of the DANN model. This layer was selected because it captures domain-invariant features that retain meaningful class-discriminative information, providing a favorable trade-off between generality and specificity. In contrast, layers in the label\_predictor branch are more specialized and prone to overfitting or retaining residual domain signal, while the domain\_classifier branch tends to suppress task-relevant structure. Therefore, feature\_extractor.dropout1 was chosen as the most interpretable and robust point in the network for downstream analysis. 

For a final model, we did not extend training until the domain classifier achieved near-zero performance, as our goal was not to completely eliminate domain information, but rather to minimize its interference with label prediction. First, clustering metrics applied to SHAP-based UMAP projections—specifically Calinski-Harabasz and Silhouette Scores—reached a plateau or declined after a certain point (Figure \ref{fig:Figure5}), indicating no further improvement in the class-relevant structure. This suggests that continuing training beyond this point may lead to diminishing returns or even degraded representation quality. Second, for effective adversarial learning, some level of domain signal must remain interpretable by the domain classifier during training—otherwise, the model would lack feedback on what features to suppress. In other words, successful confusion requires partial awareness of what is being confused. Therefore, we selected the final iteration (499 ephoc) based on when the representation was both class-informative and sufficiently domain-invariant, rather than pushing the domain classifier to complete failure.

UMAP coordinates based on these SHAP values were used to construct a k-nearest neighbor graph with n\_neighbors=400, and Leiden clustering was performed using igraph with the RBConfigurationVertexPartition method at a resolution parameter of 0.3, yielding distinct subpopulations.

To assess the coherence of these clusters, we trained a second XGBoost classifier to predict Leiden cluster membership using the original transcriptomic matrix as input. Due to memory constraints, a reduced subset of the data was used for training, allocating 30\% of the samples for this step. Importantly, the data was split using train\_test\_split with stratification based on vital status, to ensure that the survival classes were proportionally represented in both training and testing subsets—this was necessary due to the known imbalance between alive and deceased samples. Model performance was evaluated using a classification report, which provided precision, recall, and F1-score per cluster to account for imbalanced class distribution and to ensure reliable interpretation of cluster-level gene contributions.

To interpret the molecular drivers of each Leiden-defined subpopulation, we applied SHAP analysis using TreeExplainer from the SHAP library. After training the cluster classifier, SHAP values were computed for all test set samples and stratified by cluster to identify feature contributions specific to each group. SHAP summary plots were then generated to visualize the top gene-level drivers per cluster, offering interpretable insight into the molecular basis of each survival-associated subpopulation.

\subsection{SHAP Value Transformation for Layered Violin Plot Visualization}

To improve the visualization of SHAP value distributions in layered violin plots, we applied a nonlinear transformation that exaggerates the magnitude of medium and high SHAP values while preserving their original direction (sign). The transformation is defined as follows:

\[
\text{sign} = \mathrm{sign}(\mathrm{SHAP}), \quad
\text{transformed} = -1 \times \text{sign} \times \left(
\exp\left( \alpha \cdot
\frac{\log(1 + |\mathrm{SHAP}| + \varepsilon)}{\log(b)} \right) - 1
\right)
\]

Where:
\begin{itemize}
    \item \(\alpha = -2.0\) is the exaggeration strength controlling the curve's steepness,
    \item \(b\) is the logarithmic base (typically \(10\)),
    \item \(\varepsilon\) is a small constant added for numerical stability,
    \item \(\mathrm{sign}(\cdot)\) preserves the original sign of each SHAP value.
\end{itemize}

This transformation was applied element-wise to the SHAP matrix before plotting. It enhances contrast in violin plots by expanding the range of mid-level SHAP values while compressing extremely low values, making subtle but relevant patterns more visually distinguishable. Importantly, the transformation preserves the directionality of the original SHAP attributions—negative values indicate a contribution toward " alive", while positive values indicate a contribution toward "deceased".

\subsection{Identification and Visualization of Cluster-Specific Prognostic Genes}

To explore the relationship between transcriptional clusters and prognostic genes across cancer types, we conducted an unsupervised clustering of gene expression profiles from TCGA RNA-seq data. Five transcriptomic clusters (Cluster 0–4) were identified based on the unsupervised data embedding. To assess the clinical relevance of these genes, we queried the Human Protein Atlas (HPA v24.0) survival dataset, identifying genes with reported prognostic associations (favorable or unfavorable) in one or more TCGA cancer types. A parallel-categories diagram was constructed to visualize how these cluster-enriched genes distribute across tumor cohorts and survival associations.

\subsection{Implementation details}
All experiments were carried out on Google Colab Pro, utilizing a high-RAM virtual machine equipped with an NVIDIA Tesla T4 GPU (15 GB VRAM) and 51 GB of system RAM. The computational environment was based on Python 3 running on Google Compute Engine.

\section{Acknowledgments}

C-PM is a doctoral student from Programa de Doctorado en Ciencias Biomédicas, Universidad Nacional Autónoma de México (UNAM) and received fellowship to CVU 855825 from CONAHCyT, México. JO-V work was supported by CONACYT (Grant Ciencia de Frontera 2019, FORDECYT-PRONACES/425859/2020) and UNAM Posdoctoral Program DGAPA (POSDOC). OR-A thank the financial support from CONACYT (Grant Ciencia de Frontera 2019, FORDECYT-PRONACES/425859/2020), PAPIIT-UNAM (IN213824), and an internal grant from the National Institute of Genomic Medicine (INMEGEN, México).

\bibliographystyle{unsrtnat}
\bibliography{references}  






\clearpage
\section*{Supplementary Material}

\renewcommand{\thefigure}{S\arabic{figure}}
\setcounter{figure}{0}  

\begin{figure}[htbp]
    \centering
    \includegraphics[width=\linewidth, height=0.8\textheight, keepaspectratio]{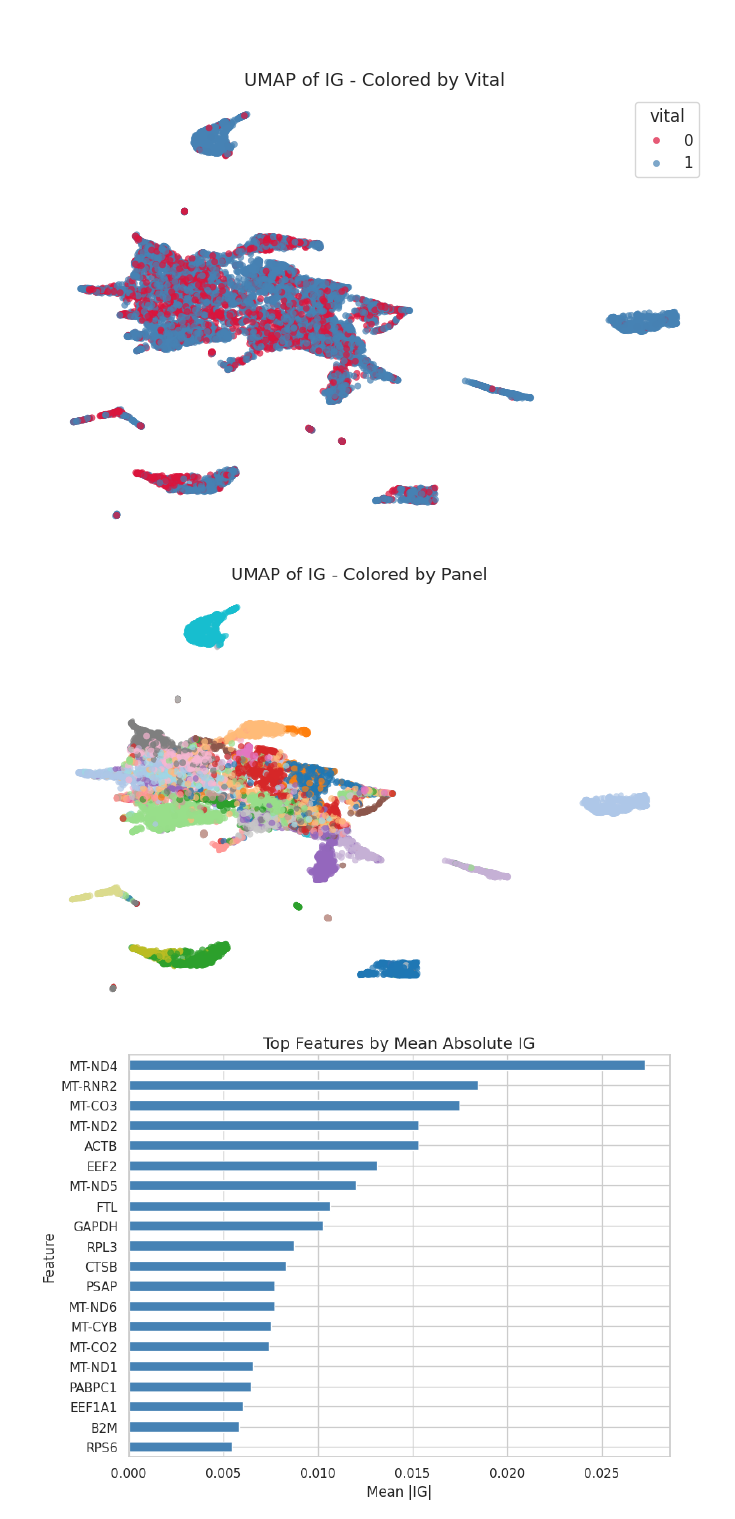}
    \caption{Integrated Gradients (IG) explanation maps for the trained DANN model, including UMAP projections of IG values colored by vital status and TCGA cancer type, and a ranking of the top transcriptomic features by mean absolute IG.}
    \label{fig:FiguraSup1}
\end{figure}

\begin{figure}[htbp]
    \centering
    \includegraphics[width=0.9\textwidth]{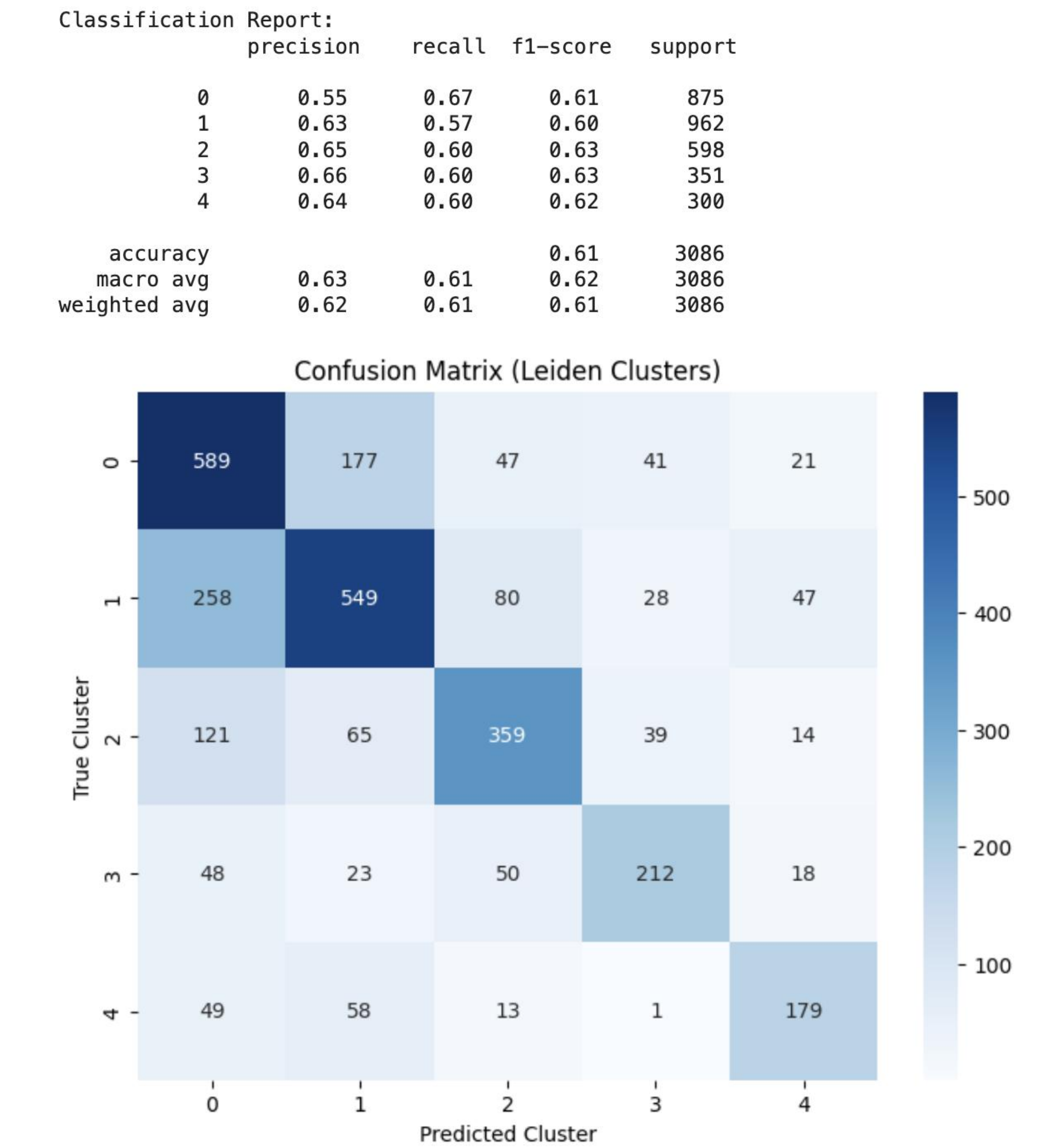}
    \caption{Classification report for Leiden clusters using the activations from the \texttt{feature\_extractor} layer at epoch 499. The report summarizes performance of a classifier trained to predict vital status based on cluster-level representations.}
    \label{fig:FiguraSup2}
\end{figure}

\end{document}